\documentclass[journal]{IEEEtran}
\IEEEoverridecommandlockouts
\usepackage{cite}
\usepackage{amsmath,amssymb,amsfonts}
\usepackage{algorithmic}
\usepackage{graphicx}
\usepackage{textcomp}
\usepackage{xcolor}
\def\BibTeX{{\rm B\kern-.05em{\sc i\kern-.025em b}\kern-.08em
    T\kern-.1667em\lower.7ex\hbox{E}\kern-.125emX}}
\usepackage{comment} 
\usepackage{color,soul}
\usepackage{float}
\usepackage{subcaption}
\usepackage{eucal}
\usepackage[normalem]{ulem}

\begin{document}

\title{Collective awareness for abnormality detection in connected autonomous vehicles \\
}

\author{\IEEEauthorblockN{Divya Thekke Kanapram (1,2), Fabio Patrone (1), Pablo Marin-Plaza (3), Mario Marchese (1), Eliane L Bodanese (2), Lucio Marcenaro (1) and David Martin Gomez (3), Carlo Regazzoni (1) \\
\IEEEauthorblockA{(1) Department of Electrical, Electronics and Telecommunication Engineering and Naval Architecture,\\ University of Genova, Italy\\(2) Centre for Intelligent Sensing, School of Electronic Engineering and Computer Science (EECS),\\ Queen Mary University of London, UK\\(3) Intelligent Systems Lab\\ Universidad Carlos III de Madrid, Spain \\
Email: \{divya.kanapram@ginevra.unige.it, f.patrone@edu.unige.it, pamarinp@ing.uc3m.es, mario.marchese@unige.it, eliane.bodanese@qmul.ac.uk, lucio.marcenaro@unige.it, dmgomez@ing.uc3m.es, Carlo.Regazzoni@unige.it \}}
}}

\maketitle

\begin{abstract}

The advancements in connected and autonomous vehicles in these times demand the availability of tools providing the agents with the capability to be aware and predict their own states and context dynamics. This paper presents a novel approach to develop an initial level of collective awareness (CA) in a network of intelligent agents. A specific collective self-awareness functionality is considered, namely agent-centered detection of abnormal situations present in the environment around any \textit{agent} in the network. Moreover, the agent should be capable of analyzing how such abnormalities can influence the future actions of each \textit{agent}. Data-driven Dynamic Bayesian Network (DBN) models learned from time series of sensory data recorded during the realization of tasks (agent network experiences) is here used for abnormality detection and prediction. A set of DBNs, each related to an \textit{agent} is used to allow the agents in the network to reach synchronously aware about possible abnormalities occurring when available models are used on a new instance of the task for which DBNs have been learned.

A Growing Neural Gas (GNG) algorithm is used to learn the nodes variables and conditional probabilities linking nodes in the DBN models;  a Markov Jump Particle Filter (MJPF) is employed for state estimation and abnormality detection in each agent using learned DBNs as filter parameters. 
Performance metrics are discussed to asses the algorithm's reliability and accuracy. The impact is also evaluated by the communication channel used by the network to share data sensed in a distributed way by each agent of the network. The IEEE 802.11p protocol standard has been considered for communication among agents. Performances of the DBN based abnormality detection models under different channel and source conditions are discussed. The effects of distances among agents and of the delays and packet losses are analyzed in different scenario categories (urban, suburban, rural).\\
Real data sets are also used acquired by autonomous vehicles performing different tasks in a controlled environment.
\end{abstract}

\section{Introduction} \label{sec:intro}
Internet of things (IoT) is a concept that connects various physical objects and allows them to exchange data over the Internet. IoT is a forefront technology which helps to reduce human efforts by enabling autonomous control capabilities and intelligence in machines and makes human life easier. IoT can produce a network of intelligent systems when combined with machine learning and signal processing techniques. IoT with smart objects has many applications in the field of surveillance \cite{ding2018amateur}, transportation \cite{guerrero2015integration}, crowd monitoring \cite{gubbi2013internet}, etc. The term Internet of Vehicles (IoV) has been defined whenever smart physical objects are vehicles \cite{gerla2014internet}. The number of vehicles is increasing exponentially as a consequence of the rapid increase in world population and the expansion of big cities. As a result, road accidents also increase dramatically by various reasons such as distracted driving, adverse weather conditions, speeding, unavailability of contextual aware data, etc. These factors highlight the need for making the objects `self-aware' and sharing these awareness data among other objects, in order to develop and enrich contextual awareness. Each object needs to be aware not only of itself but also of the other objects and conditions of the surrounding area.

Inter-connectivity and efficient communication schemes are required to develop such collective awareness among smart objects which, in case of intelligent vehicles, can help assure the safety and efficiency in driving. By making objects self-aware, each object would be able to detect abnormal situations in the environment and make appropriate decisions to avoid accidents, threats, or other dangerous actions. If we consider a network of such agents which periodically communicate the acquired information among each other, the future knowledge and actions of one agent can affect the other ones' behaviour. For this reason, reliable communication among agents is mandatory to let them successfully cooperate. However, data exchange among objects can be adversely affected by different factors, such as the distance among them, the transmission delay, and the environmental conditions.

In this work,  Dynamic Bayesian Networks (DBNs) \cite{murphy2002dynamic} have been used as data-driven models learned from sensory data for the detection of abnormalities and the prediction of future states of the agents. DBN's are hierarchical probabilistic models that make it possible to filter observations received from multiple sensors in order to understand and predict the possible object states that have generated the observations themselves. Data-driven DBN models require temporal dynamic relationships among object states that are learned from training data acquired along with the accomplishment of reference tasks by the agent.  Such learned dynamic models make it possible to analyze time-series, i.e. time sequences of observations acquired in successive experiences of the same type to estimate if they occur in statistically normal (i.e. similar) ways. A   probabilistic density function (PDF), namely the posterior of unknown states conditioned over observations is the typical recursive result obtained by a DBN filter at each time instant. A specific class of  DBN models is here used, i.e.   Switching Linear Dynamic Systems (SLDS) \cite{fox2009nonparametric} or switching models. In such models, it is possible to represent a complex non-linear dynamic behavior with a probabilistic sequential combination of linear dynamic models.  Switching random variables are used within the SLDS as higher-level hidden discrete variables one to one associated with different linear dynamic models defined in the DBN. The joint posterior of state and switching variable (superstate) of an object can be recursively estimated within a SLDS.   Markov Jump Particle Filter (MJPF) \cite{costa2006discrete} is an inference algorithm that allows estimation of posterior and prediction to be performed when a SLDS is available. Another example of Bayesian inference approach to be used for switching models,  however dealing with non-linear dynamic models,  is Rao-Blackwellized particle filter (RBPF) \cite{sarkka2007rao}.

The first objective of this work is to make it possible awareness functionalities are exhibited by each agent of a network that is performing a collaborative task. Abnormality detection is considered as a first functionality necessary to reach collective awareness in the agent's network. As abnormality detection should be performed synchronously by all agents, communication of information required to obtain it must be considered. To achieve this goal, each agent has to learn in an offline training phase, a set of switching DBN models, one for each object in the network that is executing a collaborative task. In this way, during subsequent online situations, each agent will be provided by a model of expected normal behaviors of all agents in the network, including itself. A bank of  MJPF filters will be applied to each DBN by each agent, that provides an estimation of posteriors and predictions for each agent. In addition, MJPF is provided of an abnormality detection capability, obtained by measuring the fitness of the dynamic models to current data at different levels of each DBN. Probabilistic distances between predicted priors and likelihood information inside each DBN node are used for this purpose.  The second objective is to assess the impact of realistic information exchange among objects on the abnormality detection feature of collective awareness. With a reliable and efficient communication, agents should be capable of sharing ground truth observations acquired in a distributed way by each of them with all other agents in the network. Each agent should dispatch appropriately the ground truth observations received from each remote transmitting agent to the appropriate MJPF where it can be compared with the respective state predicted by the relative DBN model to estimate the possible presence of abnormalities. In this way, each agent can estimate global abnormality conditions that can arise in any of the agents that compose the network.  Self Awareness (SA) of single agents can so become  Collective Awareness (CA) of the agent's network. Different distributed communication schemes at increasing complexity can be devised to reach such a CA. The agents could for example  either communicate all their first-person  observations to each other and then apply MJPFs for all agents to obtain abnormality estimation  or communicate and share the outputs of only their  own MJPFs using locally observed data. In this latter way, abnormality situations would be communicated, and local observations could remain private within the agent. In this paper, a first analysis is provided to indicate which communication strategy could be the best one depending on contextual parameters like the distance between the agents, the transmission delay, and the achievable data rates of the chosen communication protocol.

The main contributions of the paper can be summarized as follows:
\begin{itemize}
  \item A method is provided to learn normality models for situations providing training data series, models represented as banks of DBNs. It is shown that an agent\textquotesingle s network can use such learned models to online detect abnormal situations that occur in any of the intelligent objects of the network.  Results of specific unsupervised learning algorithms used in the training phase to estimate the DBN models, like the GNG algorithm, are provided. Also, the results coming from a specific SLDS inference method working on learned DBNs, namely a MJPF model extended to become able to detect abnormalities, are discussed.
  \item The robustness of the distributed abnormality detection feature of CA with respect to a realistic communication channel model are discussed; performances are evaluated in order to assess, on the one hand, the reliability and accuracy of abnormality detection under perfect communication hypothesis, and, on the other hand, the robustness is also analyzed of the system model against packet losses and transmission delays of the communication channel among objects.
\end{itemize}

The remainder of this paper is structured as follows. Some of the main articles and works in the literature regarding self-aware vehicular networks are reported and summarized in Section \ref{sec:stateArt}. Section \ref{sec:methodology} reports our proposed strategy for anomaly detection, describing in detail the principles exploited in the training phase, the steps included in the test phase, and how the communication among agents has been modelled. The experimental setup and the communication system are included in Section \ref{sec:experimental}, proposed anomaly detection scheme and the communication among objects are described in Section \ref{sec:results}. Conclusions and possible future work are drawn in Section \ref{sec:conclusions}.

\section{State of the art} \label{sec:stateArt}
This section describes some of the related work regarding the development of self-awareness in agents and how such agents can perform better if they are part of a network. According to the statistics, about 75 billion ``things'' will be connected to the Internet by 2025, and a larger portion will be vehicles \cite{statista}. The number of vehicles equipped with IoT technology is increasing as a consequence of the rapid growth of vehicles numbers and the spread of the IoT technology, leading to a change from the conventional Vehicle Ad-hoc Networks (VANETs) \cite{dua2014systematic} concept to the Internet of Vehicles (IoV) principle. Allowing inter-vehicles communication with the aim to make them smarter and self-aware is the main principle of IoV, where the concept of self-awareness mainly describes the cognitive capability of living entities such as humans. Self-awareness can be defined as the capability to observe itself as well as the surrounding environment, i.e. contextual awareness. If a vehicle is self-aware capable, it can be provided of models allowing it to detect abnormal situations and consequently decide emergency actions before the situation goes beyond its control. Moreover,  such important information has to be shared by communications with other vehicles in the network to make all the surrounding entities aware of the overall situation. This shared awareness is here defined as collective awareness (CA).

In \cite{baydoun2018multi}, the authors propose an approach to develop a multilevel self-awareness model by focusing on one agent. The developed self-awareness approach is learned by using multisensory data of a vehicle normally interacting in an environment. The model allows the agent to become able to detect abnormal situations present in its surrounding environment. The learning process of the self-awareness model for autonomous vehicles based on data collected from human driving is described in another work \cite{ravanbakhsh2018learning}. Other related works in this direction aim to enrich the experience of co-operative and secure driving \cite{6823640,parno2005challenges}. 

Artificial intelligence and machine learning are two multi-disciplinary concepts which are growing in interest in a number of research studies in the past few years. However, we still lack a genuine theory that explains the underlying principles and methods that would tell how to design agents that can not only understand their environment but also be conscious of what they do. Moreover, agents have to understand the purpose of their own actions to take a timely initiative beyond the already programmed goals set by humans. Another important aspect is that agents should be able to incrementally learn from their own previous and current experiences and share learned models with other agents. The issues mentioned above are not new, and researchers in various fields of science (artificial intelligence, cognitive science, neuroscience, and robotics) have already addressed the problems in organization and operation of a system capable of performing perception, action, interaction and learning up to different levels of development \cite{morin2006levels}. The term ``cognitive architectures'' is commonly used in the Cognitive Sciences, Neuroscience, and Artificial Intelligence (AI) communities to refer to propositions of system organization models designed to mimic the human mind. Most of the previous works that aimed at developing cognitive architectures did not address the issue of self-awareness. Although, some neuroscientists have considered self-awareness as an expression of consciousness \cite{hood2012self}, others propose to ground it in the robust theoretical framework of integrated information theory \cite{koch2016neural}. Firstly, it is required to investigate if and how a machine can develop self-awareness and then how it can communicate with other self-aware agents to achieve common goals. To do so, it is important to understand the concept clearly and to propose a computational model that can account for it \cite{chella2009machine,lewis2011survey}. To make the agent self aware, it is crucial to develop and integrate perceptual abilities for self-localization and environment interpretation, decision-making and deliberation, learning and self-assessment, and interaction with other agents. If we could bring all those up to the implementation level, it would be possible to make the agent self-aware, i.e., awareness of being in control of it\textquotesingle s own actions and responsible for their outcomes \cite{haggard2009experience}. In addition to that, such an integration of the results and characteristics of various subconscious deliberative processes (such as perception, action, and learning) in a shared global workspace \cite{dehaene2001towards} appears fundamental in humans to enable meta-cognitive processes such as the ability to report to oneself and to other agents about internal state, decisions and the way these decisions were made \cite{shadlen2011consciousness}. Additionally, it is vital to develop predictive models of agents \cite{seth2012interoceptive}.

To bring IoT into its next cognitive level, more sophisticated AI needs to be injected across the entire network to make it self-aware. Currently, autonomous vehicles may combine data from cameras, onboard sensors, and lidars, making them intelligent and able to learn and adapt to each possible situation. But if they are not connected, then we cannot call them smart. Uber and Tesla are self-driving vehicles, but they are not connected, and they do not cooperate with each other. The two strong technology trends, one in the mobile communications industry and the other in the automotive industry, are becoming intermixed and will provide new capabilities and functionality for future Intelligent Transport Systems (ITSs) and future driving. As the vehicles are continuously growing more aware of their environment due to the higher number of sensors they are equipped with, the amount of interactions rises, both in between vehicles and between vehicles and other road users. As a result, the significance and reliance on capable communication systems for ego-things/Machine to Machine (M2M) are becoming a key asset. On the other hand, the mobile communications industry has connected more than 5 billion people over the last 25 years, and the next step in wireless connectivity is to link all kinds of devices. According to Ericsson’s technical mobility report published in 2017, around 29 billion connected devices are forecast by 2022, of which approximately 18 billion will be related to IoT.

In \cite{baydoun2018multi}, the authors propose a method to develop a multilayered self-awareness in autonomous entities and exploit this feature to detect abnormal situations in a given context. Most of the related works \cite{han2017tdoa,ravanbakhsh2018learning} use either position-related information to make inferences or the agents are not connected to each other. In an autonomous agent, the information related to the control plays a significant role in the prediction of future states and actions of the entity. Moreover, it is imperative to establish a reliable and accurate network among the agents to allow them to communicate important awareness information in order to make the entire network fully aware of the context where the agents operate.

\section{Methodology} \label{sec:methodology}
This section first describes how the "awareness" can be modelled into the \textit{things} that can generate "ego-things". Ego-things can be defined as intelligent autonomous entities that can perceive their internal as well as external parameters and adapt themselves when they face abnormal situations. Secondly, we investigate how the network of such ego-things can establish timely and efficient communication to develop collective awareness (CA). 


The collective awareness in a network of ego-things is defined here as the capability of a set of ego-things in a network to understand whether perception-action information processing models they are provided of, are performing in a normal way. Normality is defined in a Bayesian inference sense, i.e. as the capability of dynamic models describing hidden object state characteristics as confirmed by observations of available agents ego-things' sensors. Such an ability is provided to each ego-thing in the network and concerns the whole set of agents. Communication is available in the network to exchange information necessary to detect abnormalities of all ego-things by each agent in the network. Each ego-thing can so achieve awareness not only about the fitness of its own models when predicting its own state but also about the possibility that abnormality conditions affect the actions of other cooperating agents with respect to predictions provided by their dynamic models. Such a collective awareness can trigger agents' decision systems to perform emergency routines or switching to other available modalities.

The collective awareness is based on detecting jointly and synchronously abnormal situations present in the context. It allows appropriate decisions that can be taken to maintain the stability of the entire network of systems. 

\begin{figure*}[ht]
\centering
 	\includegraphics[width = 1 \linewidth ]{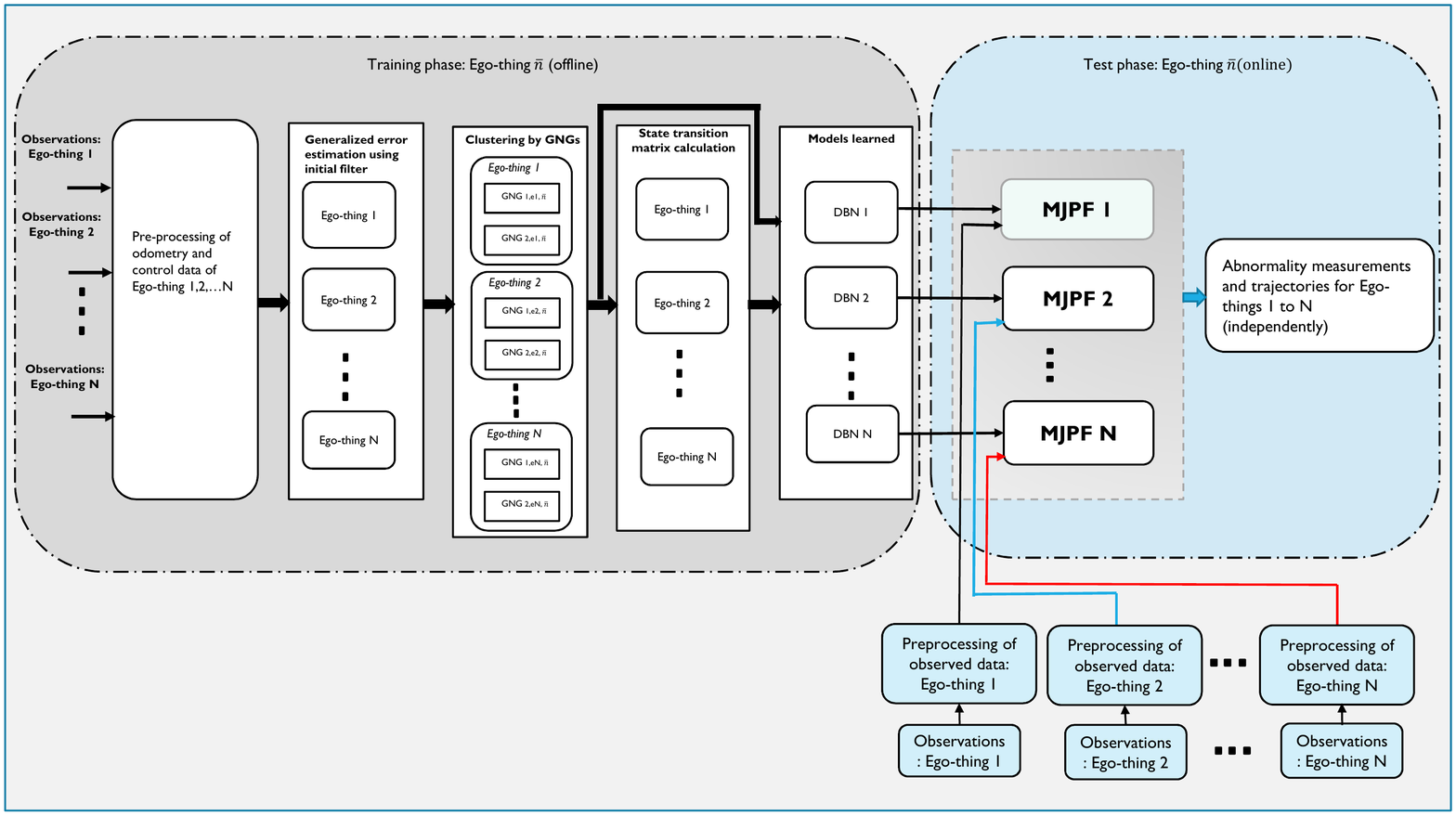}
\caption{Block diagram: training phase and test phase}
\label{fig:compactbd}
\end{figure*}

The ego-things are equipped with various sensors. The collected data from each ego-thing have been initially synchronized and then categorized into different groups. In this work, we mainly consider the data related to the control part of the ego-thing along with the trajectory data in order to develop collective awareness. The proposed method can be divided into two parts: offline training and online testing. A block diagram representation of the proposed method is shown where offline training and online processing carried on by each ego-thing in the network is shown in Fig. \ref{fig:compactbd}. During the offline training phase, each ego-thing learns probabilistic filtering models from agent sensors' dynamic data series collected while collectively performing a  reference situation task. This implies that during the training phase, all agents perform the task autonomously or in a teleoperated way. The collected data series provides information on data collected by sensors' observing esoperceptive and proprioceptive data. By assuming that observation models can remain invariant along the process (i.e. the model for estimating state likelihood from sensory observations is given and fixed), a set of dynamic models is learned, composed by a discrete vocabulary of continuous conditional probabilities functions and by a transition matrix. Such models are organized within a Dynamic Bayesian Network. Such a process is repeated for each agent, and the set of DBNs related to each agent is made available to the collective ensemble of ego things. In the online phase, each DBN is used for filtering agent sensory data within each agent. The comparison of learned dynamic prediction models with incoming observations allows each agent to estimate the level of fitness and to measure abnormality of the collective situation in a distributed way. However, to this end, communications have to be maintained to allow each agent to filter and detect abnormalities also of other ego-things in the network. Filtering is performed using a Bayesian filter appropriate for the type of DBNs learned, i.e. switching models.  Markov Jump Particle Filter (MJPF) has been here chosen as the dynamic probability models in learned DBNs are here linear and continuous Gaussian, so allowing Kalman filters to be used at a continuous level in switching models. In Figure 1, it is highlighted how such filters are here provided of the additional capability of measuring abnormalities, in addition to filtering, such capability is at the basis of CA. 

\subsection{Offline training phase} \label{subsec:training}
In the training phase, each ego-thing will learn a switching Dynamic Bayesian Network (DBN) model for itself, i.e. from the data collected by its own sensors, and one DBN model for each other ego-thing present in the network by exploiting the data generated by that ego-thing's sensors. In this work, ego-things are autonomous vehicles, and the number of vehicles is limited to two. In Fig. \ref{fig:compactbd}, the first part (grey shaded area) represents the training phase of an ego-thing, and the steps followed to learn the switching DBN models are explained below.


\subsubsection {Data pre-processing and state estimation}

First, the collected multisensory data are synchronized by using their time stamps. In this work, we considered as a case study the data sequences related to two low dimensionality sensorial data, namely odometry as representative of ego-thing's esoperceptive sense of position and steering as proprioceptive control information of the ego-thing. An initial basic generalized filter \cite{campo2017static} has been applied to the data sequence for the estimation of generalized states. The generalized state estimation of acquired data is described below.\\  
Let $Z^{e\overline{n}}_k$ be the measurements in the ego-thing $e\overline{n}$ at the time instant $k$ and $X^{e\overline{n}}_k$ be the state associated to the measurement $Z^{e\overline{n}}_k$, such that $Z^{e\overline{n}}_k = g(X^{e\overline{n}}_k) + \omega_k$. $g(\cdot)$ is the function that maps states into observations and $\omega_k$ represents the noise of the sensors. Similarly, $e\overline{n}$ will also have measurements from all the other ego-things in the network, which can be represented as $Z^{e1}_k, \dots, Z^{en}_k, \dots, Z^{eN}_k, n \in \mathcal{N}, n \neq \overline{n}$, where $N$ and $\mathcal{N}$ are the number and the set of ego-things in the network, respectively.

As explained in \cite{Friston1211,balaji2011bayesian}, including time derivatives in hidden object states allows dynamic probabilistic flow models describing ego-thing states to be one to one related with descriptors of motion laws coming from the mechanical statistic, i.e.Lagrangian. Such flow models are represented in the moving reference system of each ego-thing so allowing data series to be described as relative not only to the estimated state of the object but also to how such a state is instantaneously changing.  The generalized state of $e\overline{n}$ by considering only itself can be defined as:
\begin{equation}\label{eq1}
\boldsymbol{X}^{e\overline{n}}_k = [X^{e\overline{n}}_k \hspace{2mm}  \dot{X}^{e\overline{n}}_k \hspace{2mm} \ddot{X}^{e\overline{n}}_k  \hspace{0.5mm} \cdots \hspace{0.5mm} X^{(L)e\overline{n}}_k]^\intercal,
\end{equation}
where $(L)$ indexes the $L$-th time derivative of the state.

The $l$-th time derivative in $e\overline{n}$ at the time $k$ by considering only itself can be approximated as: 

\begin{equation}\label{eq2}
X^{(l)e\overline{n}}_k = \frac{X^{(l-1)e\overline{n}}_k - X^{(l-1)e\overline{n}}_{k-1}}{\Delta k},
\end{equation}

where $X^{(0)e\overline{n}}_k = X^{e\overline{n}}_k$ and $\Delta k$ is the uniform sampling time for all multisensory data.

The generalized state of $e\overline{n}$ by considering all the ego-things in the network can be written as:\\

\begin{equation}\label{eq2b}
\boldsymbol{C}^{e\overline{n}}_k = [\boldsymbol{X}^{e1}_k \hspace{2mm}  \boldsymbol{X}^{e2}_k \hspace{2mm} \boldsymbol{X}^{e3}_k  \hspace{0.5mm} \cdots \hspace{0.5mm} \boldsymbol{X}^{eN}_k]^\intercal,
\end{equation}

\subsubsection{Clustering by GNGs}
When a data series is available, a generative filter capable of generating other instances provided of the same statistical properties as well as to predict future states has to be learned. Generative filters here used are hierarchical switching 2-Time Slice DBNs (2T-DBN) \cite{Koller2001SamplingIF}. This generative filter, as shown in Fig. \ref{fig:CDBN} is composed of hidden states at continuous and discrete levels. Generalized states are here used at the continuous level.  Discrete hidden states are hierarchically higher and represent switching variables. For each value of such random variables, a different dynamic model has to be learned at the continuous level capable of predicting in a different way dynamics of states. This type of DBNs is capable of representing non-linear dynamic models by using a set of linear dynamic models. In order to learn such DBNs,  the vocabulary of switching variables and the associated set of dynamic linear models must be learned from data. To this end, having used generalized states is particularly useful. In fact, a technique can be used as in \cite{campo2017static} that allows defining an initial basic generalized filter \cite{Friston1211} that operates on data series to produce an estimation of the dynamic model that should be associated to each state sparse point obtained by filtering the data sequence. Such a technique consists of an initial filter based on a single value switching variable; such a value corresponds to a unique dynamic model that assumes no state change should be associated with values in the data series. When a data series violates this assumption, obtained derivatives of state correspond to errors with respect to such a hypothesis. Errors can be clustered to define a set of state-dependent linear dynamic models characterizing the state as varying according to average derivatives and their covariances. Jointly clustering in an unsupervised way states and errors allow one to obtain a vocabulary of regions. Each region is characterized by a compact part of the state space and by a compact subspace of the derivative state space. The average  state derivative in the region subspace defines a different filter for each  compact state subspace. Here we used an unsupervised clustering approach, the Growing Neural Gas (GNG) \cite{Fritzke:1994:GNG:2998687.2998765} method to obtain regions from generalized errors produced as outputs by the initial filter, i.e. sequences of coupled state estimations and errors. GNG clusters correspond to coupled compact regions of state points and errors. Derivative errors cluster  encode the description of the expected dynamics that caused the data series to vary instead of following the hypothesis of the initial filter. Different ways of changing are coded as behaviors that have been found  in a  corresponding compact state region. Compact state region represents switching variables in the hierarchical DBN. GNG is not the unique possible clustering  algorithm that could have been employed. K-means clustering \cite{macqueen1967some}, Self Organizing Map (SOM) \cite{kohonen1982self}, Neural Gas (NG) \cite{fritzke1995growing},etc are other possible choices. In comparison to K-means and SOM, NG converges faster and also it has other advantages. The Growing Neural Gas (GNG) algorithm is an improved version of the NG algorithm. In comparison to NG, it does not need any dynamically modifiable parameters. The Growing Neural Gas algorithm extends the Neural Gas algorithm by adding a local error measure for each node. A second addition here used, first proposed by \cite{fritzke1997self}, is the utility measure. By considering all the aforementioned advantages, we chose to use the GNG algorithm in this work.

The output of  GNG consists of a set of clusters defined as nodes.  In the proposed approach a separate GNG clustering is applied to states and derivatives obtained from the initial filter.  Each node groups a subset of samples of states or derivatives that have a low distance with respect to the centre of mass of the region associated with the node. Iterative presentation of the same set of samples allows reorganization of nodes averages until convergence is reached.  Nodes produced by  GNG can be seen as a set of \textit{letters} forming a vocabulary. A different vocabulary is formed for GNGs working on state and derivative samples produced by the initial filter.  Nodes associated with the GNG working on derivatives form a vocabulary of dynamic linear models. The flow model of each dynamic model is defined by the centre of mass of the error in the respective node.  On the other side nodes associated with the GNG working on states defines a vocabulary of regions, i.e. switching variables of the state space.   The set of nodes produced as output by  GNG $l$, i.e. related to the $l$-th derivative order, of the ego-thing $e\overline{n}$ can be defined as:

\begin{equation}\label{eq3}
\boldsymbol{V}^{(l)e\overline{n}} = \{V^{(l)e\overline{n}}_{1}, V^{(l)e\overline{n}}_{2}, \dots, V^{(l)e\overline{n}}_{G^{(l)e\overline{n}}}\},
\end{equation}

where $G^{(l)e\overline{n}}$ is the set  of nodes of the GNG $l$ related to ego-thing $e\overline{n}$'s $l$-th derivative of the state. 

 $V^{(l)e\overline{n}}_{n}$ defines the node, and it is considered as a Gaussian random variable whose mean value is the average of samples and whose size corresponds to the variance of the samples themselves. $\boldsymbol{V}^{(l)e\overline{n}}$ can be seen as a \textit{vocabulary} of order $l$ composed by the relative nodes. The switching variables at the highest level of the  DBN model learned by GNGs so computed in Fig.\ref{fig:CDBN} is the switching variable. Such a variable assumes values from the vocabulary learned by GNG working at the state level $l=0 $. Each region can so be seen as a switching variable:  Each value of the variable indexes a region in the continuous state space corresponding to a Gaussian having as mean and covariance associated with the node. The dynamic model associated with that region is found by identifying a letter in the vocabulary of higher derivatives GNG nodes that specifies the velocity and higher-order generalized coordinates of a set of dynamic models that can be associated with the state region.

The compact regions of the derivative state space form a  \textit{vocabulary}  composed of symbols associated with different dynamics of generalized states $\boldsymbol{X}^{e\overline{n}}_k$. In this work, generalized states include only states and their  first order derivatives such as 
$\boldsymbol{X}^{e\overline{n}}_k = [X^{e\overline{n}}_k \hspace{2mm}  \dot{X}^{e\overline{n}}_k]^\intercal$.
A generic element (letter) of the vocabulary describing clusters of state derivatives can be associated with an equation of a dynamic model to be used by a linear filter. Such a model can be written as 

\begin{equation}\label{eq5}
\boldsymbol{X}^{e\overline{n}}_{k+1} = A\boldsymbol{X}^{e\overline{n}}_{k} + BU_k +  w_k
\end{equation}

where
$$A = 
\begin{bmatrix}
    I_{j}   & 0_{j,j}   \\
    0_{j,j} & 0_{j,j}   
\end{bmatrix} \hspace{1mm} ; \hspace{1mm} B = 
\begin{bmatrix}
    0_{j,j}   \\
    I_{j}\Delta k    
\end{bmatrix}$$

The variable $j$ is related to the dimensionality of the state vector for data under consideration. $I_j$ is an identity matrix of dimension $j$. $0_{j,j}$ is a zero $j \times j$ matrix. $w_{k} \sim \mathcal{N}(0,\sigma)$, encodes the noise produced by the system. $U_{k}$ is a control vector that is defined from the average derivative of states obtained by GNG within a dynamic model region. The dynamic model to be chosen is the one of the state regions to which $\boldsymbol{X}^{e\overline{n}}_{k}$  belongs. A different dynamic model can be associated to different letters describing the same state space region. 

By combining letters of nodes produced by GNG working on different derivatives, it is possible to obtain a set of \textit{words} which define discrete states combined with dynamic models, so providing a semantic vocabulary whose elements combine centroids of different derivative orders.  Such words computed at ego-thing $e\overline{n}$ are defined as:

\begin{equation}\label{eq4}
\boldsymbol{W}^{e\overline{n}} = \{\varphi^{e\overline{n}}, \dot{\varphi}^{e\overline{n}}, \dots,  \varphi^{(L)e\overline{n}}\},
\end{equation}

where $\varphi^{e\overline{n}} \in \boldsymbol{V}^{(l)e\overline{n}}$. $\boldsymbol{W}^{e\overline{n}}$ contains all possible combinations of switching variables and dynamic models. The switching variable acts as a  variable at a higher hierarchical level that explains the states from a semantic viewpoint. The discrete switching variables (i.e., letters and words) of the learned  DBN model shown in the pink shaded area in Fig. \ref{fig:CDBN}.

\subsubsection{Estimation of state transition}

The vocabularies are learned by applying initial filters and GNG clustering to each ego-thing $e\overline{n}$ sensory data acquired along a cooperative task performed with other ego-things. For example,  in  \textit{scenario 1} (refer section \ref{sec:experimental}) a cooperative driving task of two autonomous cars is considered. In order to allow each ego-thing to develop models that consider time evolution not only at continuous level but also as probabilistic transitions among \textit{words} in the learned  \textit{vocabularies}, timestamps are assumed to be provided to data series and transition models to be used at the discrete level of DBNs are estimated. Such transition models allow switching variables to be predicted probabilistically at each moment by the DBN. Moreover, as the DBNs estimate at each time, a joint posterior probability over switching models and continuous states, the predictions provided by the transition model can be used as a source to obtain a further measurement of semantic abnormality.  In particular, if predicted words do not match with evidence supported by observations of one agent, then such an agent can occur in a semantic abnormality.\\

The probabilistic transition matrix has been estimated from the data sequence by considering the transitions in time and such a matrix can tell the mapping of the variables in discrete space (i.e., word space). In other words, it can tell the probability of transition from word $W^{en}_{k}$ at time instance $k$ to the word $W^{en}_{k+1}$ in next time instance $k+1$ shown in Fig. \ref{fig:CDBN}. We use this information for the prediction purpose at the word level.

\subsubsection{DBN model for all the agents}
All the previous steps are the step by step learning process of the switching DBN models. Each ego-thing learns a total number of $N$ switching DBN models in order to predict the future states of each entity in continuous as well as discrete levels. The set of DBNs learned by each ego-thing $ei$ and $ej$ is the same for each other ego-thing in the network, and can be written as: 

\begin{equation}\label{eqDBN}
\small
 \boldsymbol{DBN}^{ei} = \{DBN^{e1}, \cdots, DBN^{eN}\} = \boldsymbol{DBN}^{ej}, \forall i,j \in \mathcal{N}
\end{equation}

The number of DBNs learned can be represented as shown in Fig.\ref{fig:CDBN}. In each DBN, there are three levels such as measurements, continuous and discrete levels. The arrows and links (coloured in black) of such DBN are learned based on the \textit{Scenario 1} task (see section \ref{sec:experimental}). Moreover, the red and blue dotted arrows represent the influence of one ego-thing\textquotesingle s action to the future states of the other ego-things in the network. These dotted arrows represent how one ego-things action can be influenced by the future actions of the other ego-things in the network and vice versa. 

\subsection{Online testing}

\begin{figure*}[ht]
\centering
\includegraphics[width = 1 \linewidth]{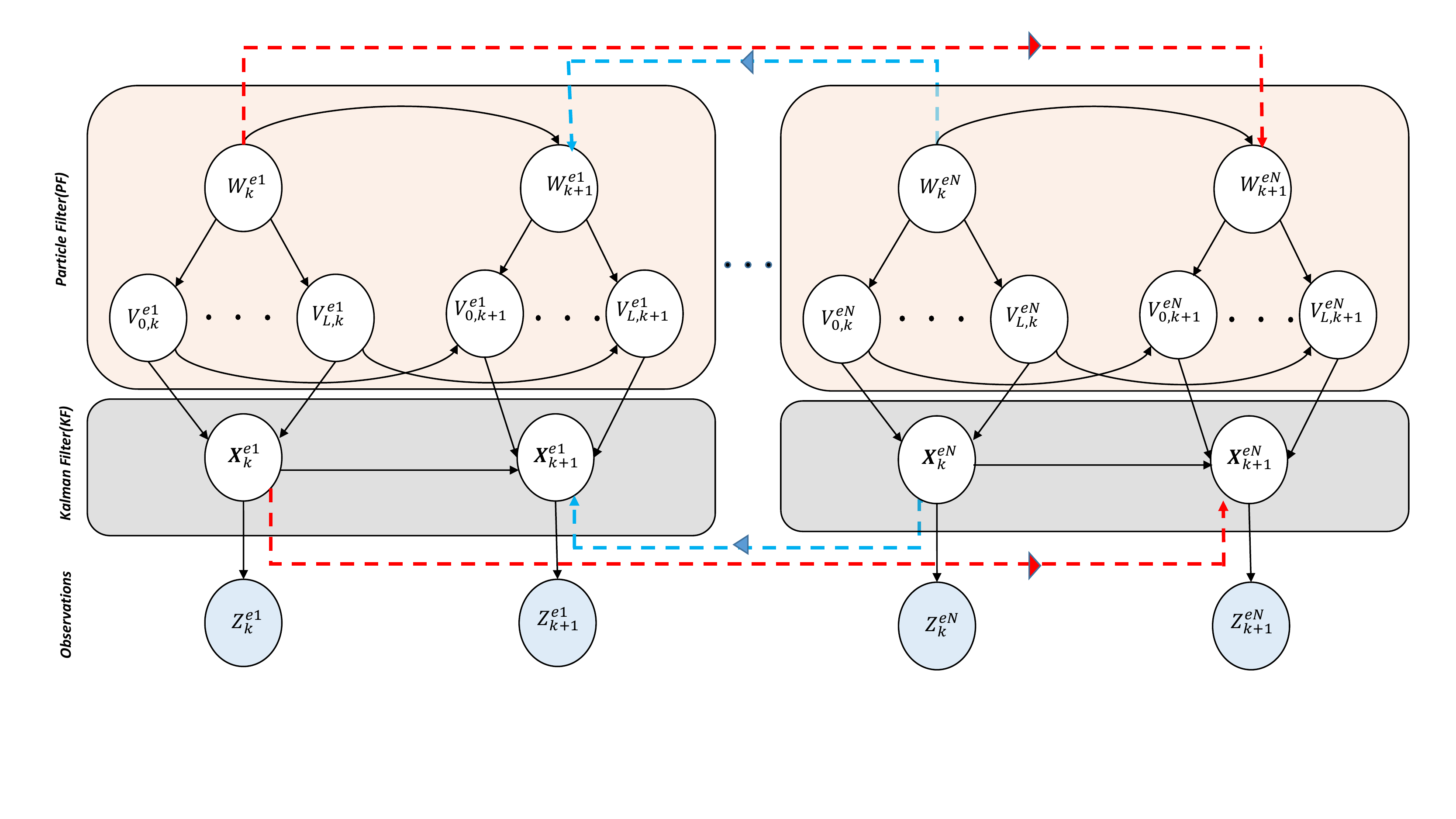}
\caption{Collective Dynamic Bayesian Network (CDBN) model}
\label{fig:CDBN}
\end{figure*}

In Fig. \ref{fig:compactbd}, the second part (shaded in blue) shows the block diagram representation of the online test phase. In this phase, we have proposed to apply a dynamic switching model called Markov Jump Particle Filter (MJPF) \cite{baydoun2018learning,8767204} to make inferences on the DBN models learned in the training phase as shown in Fig. \ref{fig:CDBN}. MJPF is a Bayesian filter with a Kalman Filter (KF) is associated with each particle. In MJPF we use Kalman Filter (KF) \cite{welch1995introduction} in state space (grey shaded area) and Particle Filter (PF) \cite{210672} in a higher hierarchical level called \textit{word} level (pink shaded area) in Fig. \ref{fig:CDBN}. The blue and red arrows in Fig. \ref{fig:CDBN} depict the information exchange between two ego-things, and, as a consequence, two DBN models. Those arrows tell how the future states of one ego-thing can influence the next states of the other one.

\subsubsection {Estimation of future states}
The MJPF is able to predict and estimate discrete and continuous states of the ego-things. In addition to that, it produces another information i.e abnormality measurements. \\
The data sequence (experience) never seen in the online training step is pre-processed and given as input to the MJPF applied on learned DBN models. The output of each MJPF is the estimation of future states of the associated ego-thing along with the probabilistic and spatial abnormality measurements.

MJPF uses PF for discrete variables, here corresponding to word variables;  each particle used to approximate the joint posterior in MJPF is augmented with an associated continuous random variable characterized by a Gaussian probability. In our case the dynamic model describing changes of  the continuous variable associated to a given word value is represented as in Eq. \ref{eq5}, while transition model is used as dynamic model at word level In the prediction step of the MJPF,  a SIR  \cite{yang2005particle} PF approach is used  to predict new candidate particles at next step using transition model at discrete level.  Each particle is enriched also of a Gaussian prediction of the continuous associated variable. This is done as in Kalman Filter, being dynamic models and observation models linear and variables Gaussian. Each predicted particle is so characterized as a word of given value with an associated prior probability at the continuous level. In the update step, the ground truth observations (belonging to each ego-thing) are used to first update the prior at continuous level, so obtaining the new posterior, and then providing a new weight to the particle word, based on the evidence that such a posterior provides  to the specific word itself. In our approach these traditional MJPF are enriched by the computation of abnormality measurements as described in the next section, to allow agents to be aware of the fitness of their dynamic models to the observed sequences.

The posterior probability density function of MJPF belongs to ego-thing ${e\overline{n}}$ can be written as:
\begin{equation}\label{eqmjpf}
 p({W_k}^{e\overline{n}},\boldsymbol{X_k}^{e\overline{n}}/ {Z_k}^{e\overline{n}}) = p(\boldsymbol{X_k}^{e\overline{n}}/{W_k}^{e\overline{n}},{Z_k}^{e\overline{n}})p({W_k}^{e\overline{n}}/{Z_k}^{e\overline{n}})
\end{equation}

where ${W_k}^{e\overline{n}}$ is the word in the higher hierarchical level and  $\boldsymbol{X_k}^{e\overline{n}}$ is the continuous state in the state space belongs to ego-thing ${e\overline{n}}$ at time instant $k$.\\
As stated above, a different Kalman Filter is associated with each particle ${W_k}^{\ast}$ and is different for each discrete zone (cluster). The Eq. \ref{eqmjpf} shows the link between the discrete state (i.e. words) and continuous state estimation. The KF associated to particle ${W_k}^{\ast}$ is used to estimate the prediction on the continuous state $\boldsymbol{X_k}^{e\overline{n}}$ and to estimate $p(\boldsymbol{X_k}^{e\overline{n}}/{W_k}^{e\overline{n}},{Z_k}^{e\overline{n}})$.\\

As explained before, each ego-thing has its own switching model as well as the model of other ego-things. At each instant, the ego-thing predicts its own future states and future states of the other ego-things by the learned switching DBN models. By receiving the ground truth observations, the ego-thing can match with the predicted states and detect if any anomalies present. The observations from other ego-things can be received through the established wireless channel with a certain delay and loss. By making efficient communication between the ego-things, we can develop the collective awareness in the entire network of ego-things. Such collective awareness can tell if any abnormal situations happen anywhere in the network. Moreover, the collective DBN models can handle the uncertainty of the environment and the variability of observations.

\subsubsection{Abnormality detection}

Posterior probability estimation in MJPF is here enriched with computation of an additional information useful for self-awareness of individual ego-things. i.e., abnormality measurements. Such information is estimated to instantaneously allow each ego-thing to measure how well the learned models fit the currently observed sequence. To estimate the abnormality of a sequence, a  statistical distance metric is here proposed that estimates the distance between predictions performed within MJPF at discrete and continuous levels and the sensory observations produced along an ego-thing experience. In this work, Hellinger Distance (HD) \cite{beran1977minimum} is proposed as the metric to evaluate sequence  abnormality.

Some important statistical distances include Bhattacharya distance \cite{bhattacharyya1943measure}, Hellinger distance \cite{beran1977minimum}, Total variation distance \cite{verdu2014total}, etc. The Bhattacharyya distance measures the similarity of two probability distributions. It is closely related to the Bhattacharyya coefficient, which is a measure of the amount of overlap between two statistical samples or populations. Similarly, the Hellinger distance (closely related to, although different from, the Bhattacharyya distance) is used to quantify the similarity between two probability distributions. The Hellinger distance is defined between vectors having only positive or zero elements \cite{abdi2007encyclopedia}. The datasets in this work are normalized, so the values vary between zero and one; there aren’t any negative values. For this reason, Hellinger distance is more appropriate than using other distance metrics as an abnormality measure. The works in \cite{8767204} and \cite{lourenzutti2014hellinger} used Hellinger distance as an abnormality measurement.


In this work, Hellinger distance is used as an abnormality measurement between predicted generalized states and observation evidence.

The Hellinger distance related to the Ego-thing $e\overline{n}$ can be written as:

\begin{equation}\label{eq6}
\theta_k^{e\overline{n}} = \sqrt{1 - \lambda_k^{e\overline{n}}},
\end{equation}

where $\lambda_k^{e\overline{n}}$ is defined as the Bhattacharyya coefficient \cite{Bhattacharyya1943}, such that:

\begin{equation}\label{eq7}
\lambda_k^{e\overline{n}} = \int \sqrt{p(\boldsymbol{X}_k^{e\overline{n}} |\boldsymbol{X}_{k-1}^{e\overline{n}} ) p(Z_k^{e\overline{n}} |\boldsymbol{X}_k^{e\overline{n}} )} \hspace{1mm} \mathrm{d}\boldsymbol{X}_k^{e\overline{n}}.
\end{equation}

The variable $\theta_k^m \in [0,1]$, where values close to $0$ indicate that ground truth observations match with predictions; whereas values close to $1$ show the presence of an abnormality.

Once detected abnormal situations, the ego-thing has to take appropriate actions either by stopping itself or reducing the speed, etc. However, the decision making part is not included in this work.
\subsection{Evaluating the model performance after the packet loss}
Each ego-thing has its own model for prediction of the future states and the ground truth observations received from the sensors to check whether if any anomalies present in the environment around it. At the same time, the models for other ego-things can predict the future states and receive ground truth observations from the corresponding ego-things with a certain delay and loss in transmission. The DBN model of each agent is updated sequentially (with a certain delay) using the shared information. The delay and the loss depend on various factors such as the distance between the ego-things, the employed communication protocol, modulation scheme, and frequency, scenario conditions (urban, rural, ...), etc.

To check the model performance in predicting abnormal situations the true positive rate ($TPR$) and false positive rate ($FPR$) are calculated to build a set of Receiver Operating Characteristic ($ROC$) curves  \cite{hanley1982meaning} corresponds to different $K$ factor values \cite{6802414}. The ROC curves plot $TPR$ and $FPR$ at different thresholds, where:

\begin{equation}\label{eq8}
TPR = \frac{TP}{TP + FN} \hspace{1mm}; \hspace{4mm} FPR = \frac{FP}{FP + TN}
\end{equation}

The true positive ($TP$) is defined as the number of times where abnormalities are correctly identified. False negative ($FN$) consists of the times that abnormalities are classified incorrectly. Accordingly, false positive ($FP$) are the times where anomalies are wrongly assigned to normal samples, and true negative ($TN$) represents the times where normal samples are correctly identified. In this work, mainly considered two parameters of ROC for measuring the performance of the model before and after the transmission loss are: \textit{(i)} the area under the curve ($AUC$) of the ROC curves, which quantifies the performance of the DBNs' abnormal detection at several thresholds; \textit{(ii)} the accuracy ($ACC$) measurement, which is defined as follows:

\begin{equation}\label{eq9}
ACC = \frac{TP + TN}{TP +TN +FP + FN},
\end{equation}

\subsubsection{Communications among ego-things}
In the training phase, we assumed that each ego-thing has available all the required data describing all the other ego-things' status. However, in a real scenario, data exchange among ego-things through a wireless mean has to be considered. Different variables affect communication performance over time. They are mainly related to:

\begin{itemize}
\item objects' movement, such as object's velocity, acceleration, and moving direction;
\item environment where the objects are located, such as urban or rural scenario, presence of obstacles, Line of Sight (LoS) or Non-LoS (NLoS) conditions;
\item chosen communication parameters, such as employed communication protocol and modulation, exploited frequency band, achievable data rate, transmission power, and received signal strength.
\end{itemize}

The channel among ego-things has to be properly modelled in order to consider all the effects which can affect the obtained performance, such as scattering, diffraction, reflection, shadowing, and fading.

The effects on the wireless channel are addressed by large scale and small scale channel models. Large scale models cover effects such as path loss and the effects of the propagation environment over large distances. Small scale models, on the contrary, describe the behaviour in the time domain, taking into account the fast fading effects, i.e. multipath propagation. Large and small scale models are combined to realistically shape channel behaviours. Received power $P_r$ is composed of the transmit power $P_t$, the large scale effects, i.e. path loss $P_L$, and the small scale effects $\zeta$:

\begin{equation}\label{eq:power}
P_r =  P_tP_L\zeta
\end{equation}

The path loss is the radio attenuation due to the communication mean. It is mainly affected by the communication frequency $f$ and the distance $d$ between source and destination. It can be computed as:

\begin{equation}\label{eq:loss}
P_L =  \bigg(\dfrac{\lambda^2}{(4\pi)^2d^\alpha}\bigg)G_RG_T
\end{equation}

where $\lambda=2 \pi f$, $\alpha$ is the attenuation factor, $G_R$ and $G_T$ are the reception and transmission antenna gains, respectively.

The presence of objects and obstacles in the environment originates multiple copies of each transmitted signal, which can strengthen (if $\zeta > 1$) or weaken (if $\zeta < 1$) the original signal. This effect is called multipath fading and can be modelled as a Rayleigh, Rician, or Nakagami distribution.

Considering the current state-of-the-art, we focus on a Rician channel model based on a Rice distribution when LoS is present. Rice distribution can be expressed with parameters $K$ and $P_r$, which are the Rician $K$ factor and the received power, respectively, or as a function of $\rho$ and $\sigma$, which are field strength of the LoS component and the field strength of scattered components, respectively.

The Rice distribution is:
\begin{equation}\label{eq:rician}
p_Z(z) = \dfrac{z}{\sigma^2}exp\bigg(\dfrac{-z^2-\rho^2}{2\sigma^2}\bigg)I_0\bigg(\dfrac{z\rho}{\sigma^2}\bigg)
\end{equation}

where $z \geq 0$, $\rho$ and $\sigma$ are the signal strength of the dominant and of the scattered paths, respectively. Therefore, $\rho^2$ and $2\sigma^2$ are the average power of the LoS and NLoS multipath components, respectively. As the direct wave weakens, the Rice distribution becomes Rayleigh.

Rician $K$ factor is defined as :

\begin{equation}\label{eq:kfactor}
K= \dfrac{\rho^2}{2\sigma^2_0}
\end{equation}

It expresses the ratio between the dominant component to scattered waves. The stronger the line of sight component, the greater the $K$ factor. In this way, the Rice distribution in eq. (\ref{eq:rician}) can be expressed in terms of linear $K$ factor as:

\begin{equation}\label{eq:eqkfactor}
\begin{aligned}
p_Z(z) = & \dfrac{2z(K+1)}{P_r}exp\bigg(-K-\dfrac{(K+1)z^2}{P_r}\bigg) \cdot \\
    & \cdot I_0\Bigg(2z\sqrt{\dfrac{K(K+1)}{P_r}}\Bigg)
\end{aligned}
\end{equation}

where $I_0$ is the modified Bessel function of first kind and zero order \cite{goldsmith2005wireless}. 
When $K \rightarrow \infty$, the Rice distribution tends to a Gaussian one, and when $K \rightarrow 0$, i.e. in case no dominant direct path exists ($\rho = 0$), the Rician fading reduces to a Rayleigh fading defined by:

\begin{equation}\label{eq:rayleight}
p_Z(z) = \dfrac{z}{\sigma^2}exp(-\dfrac{z^2}{2\sigma^2})
\end{equation}

A more general fading distribution was developed whose parameters can be adjusted to fit empirical measurements. This distribution is called the Nakagami and is given by:

\begin{equation}\label{eq:nakagami}
p_Z(z) = \dfrac{2m^mx^{2m-1}}{\Gamma(m)P^m_r}exp(\dfrac{-mz^2}{P_r})
\end{equation}

The Nakagami distribution is parametrized by $P_r$ and the fading parameter $m$. For $m=1$ it becomes Rayleigh fading, instead for $m = \dfrac{(K+1)^2}{2K+1}$ the distribution is approximately Rician with parameter $K$.

\section{Experimental set up} \label{sec:experimental}
 
\begin{figure}[ht]
	\begin{subfigure}[t]{0.25\textwidth}
		\centering
		\includegraphics[width=4cm,height=3cm]{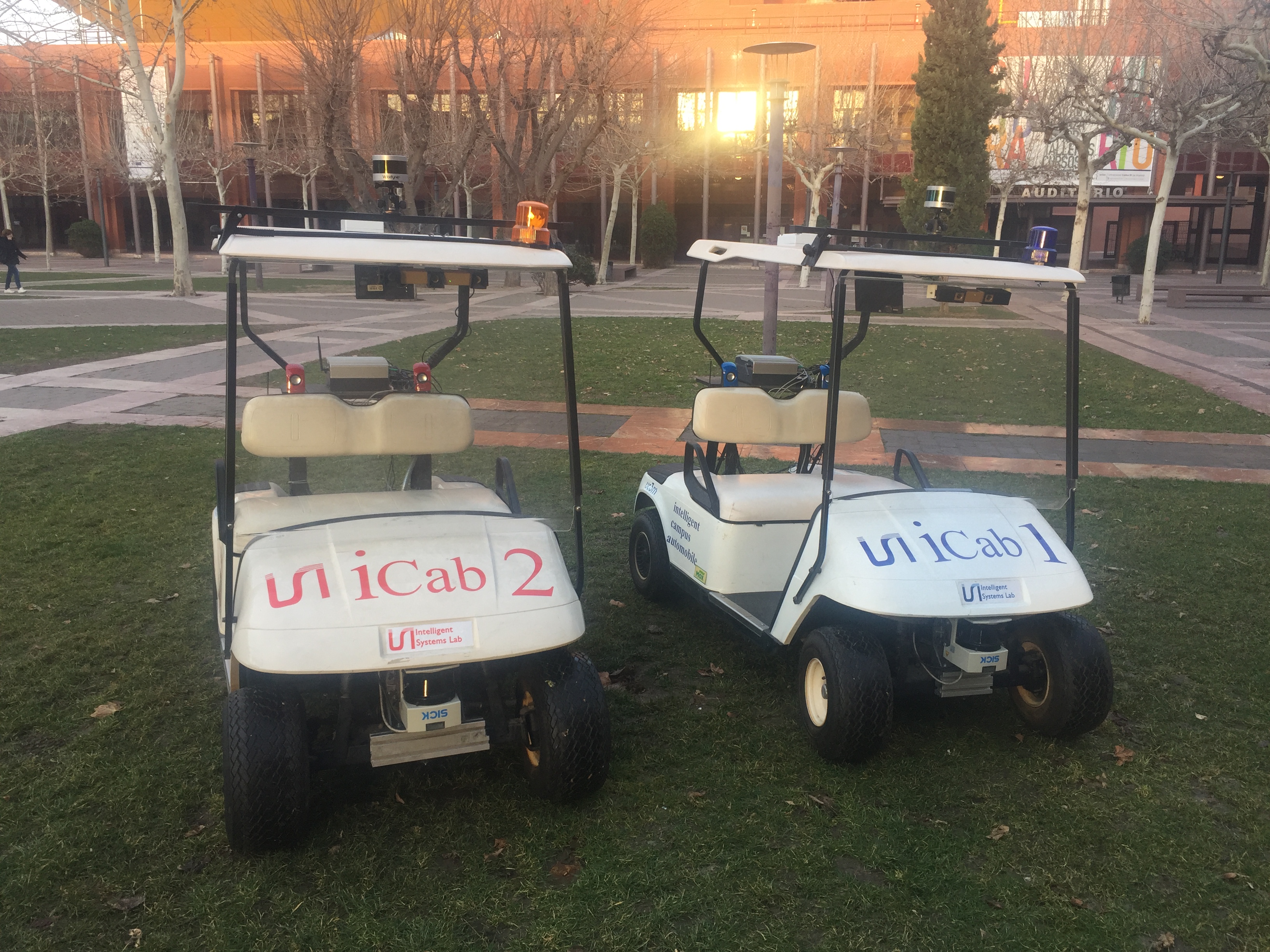}
		\caption{iCab platforms}
		\label{fig:iCab}
	\end{subfigure}%
	\begin{subfigure}[t]{0.25\textwidth}
		\centering
		\includegraphics[width=4cm,height=3cm]{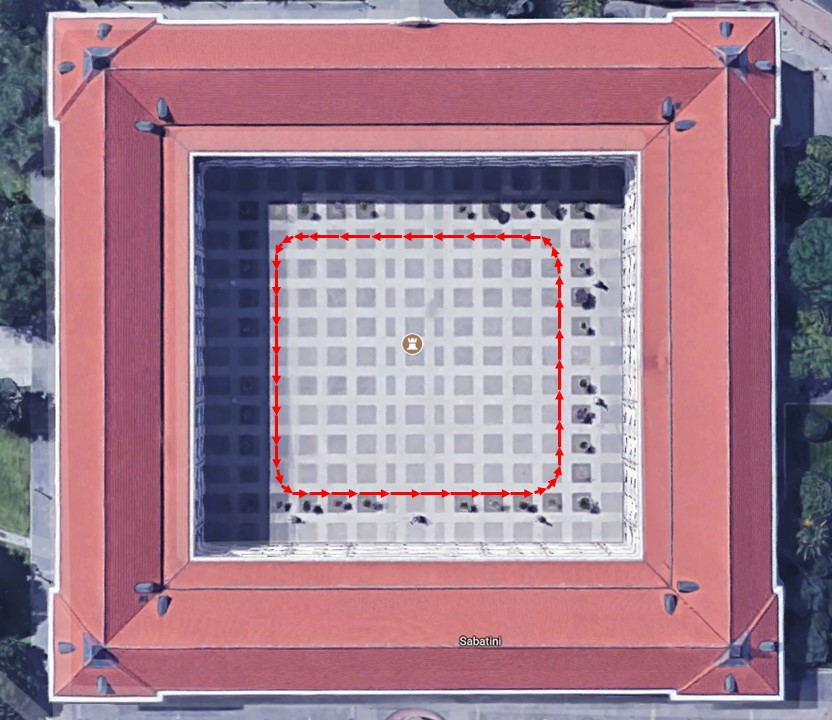}
		\caption{Testing environment}
		\label{fig:Environment}
	\end{subfigure}
	\caption{The vehicles and the environment used for the experiments.}
	\label{fig:Perimeter monitoring}
\end{figure}

The scenario considered to validate the proposed methodology consists of two intelligent vehicles called iCab (Intelligent Campus Automobile), shown in Fig. \ref{fig:iCab}, with the capabilities of autonomous driving \cite{marin2018global}. The vehicles are equipped with different sensors such as one lidar, a stereo camera, and encoders. Information about the control of the vehicles, i.e. steering angle ($s$) and power ($p$), along with the odometry data ($x$ and $y$ positions) are the data exchanged during the operative process. After collecting the data sets, a synchronization operation is performed in order to perfectly synchronize the collected data sets. The two vehicles follow the same movement trace, shown in Fig. \ref{fig:Environment}, keeping their position one after the other. For this reason, the iCab1 vehicle is called \textit{leader} and the iCab2 vehicle is called \textit{follower}. The dimension of the movement trace in the testing environment is $38 m X 33 m $. 

\begin{figure}[ht]
\centering
 	\includegraphics[width=7cm,height=7cm]{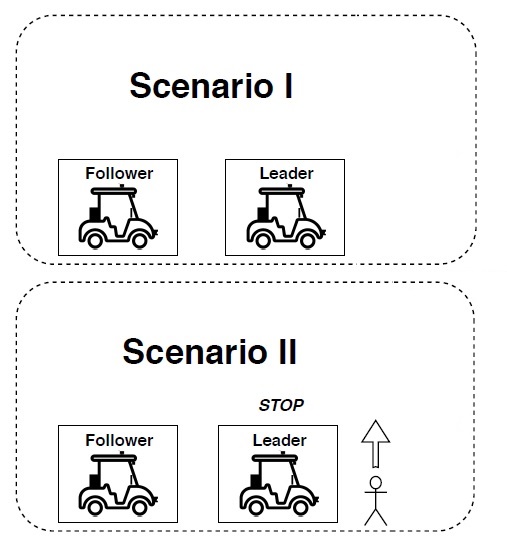}
\caption{Scenarios considered for the anomaly detection test}
\label{fig:scenario}
\end{figure}

To test the anomaly detection model, we used the data sets from the vehicles while they perform two different actions in the same test environment. The scenarios are:
 
\begin{itemize}
    \item \label{1} Scenario I \textit{Perimeter monitoring}: iCab vehicles perform platooning operation in a closed environment, as shown in Fig. \ref{fig:scenario}. In total, four laps (i.e., the platooning operation has been performed four times one after the other) has been performed and collected the data. The follower vehicle mimics the actions of the leader vehicle. This is the scenario used in the training phase to learn the switching DBN models. 
    \item \label{2} Scenario II \textit{Emergency stop}: while both vehicles are moving along the rectangular trajectory of the perimeter monitoring task, a pedestrian suddenly crosses in front of the \textit{leader} vehicle. As soon as the \textit{leader} detects the presence of the pedestrian, the vehicle automatically executes an emergency brake and waits until the pedestrian crosses and then continues the task. At the same time, the follower detects the anomaly in the state of the leader and it also performs a stop operation until the leader starts its movement again. The datasets from this scenario have been used to test the switching DBN models learned in the training phase. We have used about 30\% of the datasets to test the learned DBN models.
\end{itemize}

\begin{figure}[ht]
\centering
 	\includegraphics[width=9cm,height=7cm]{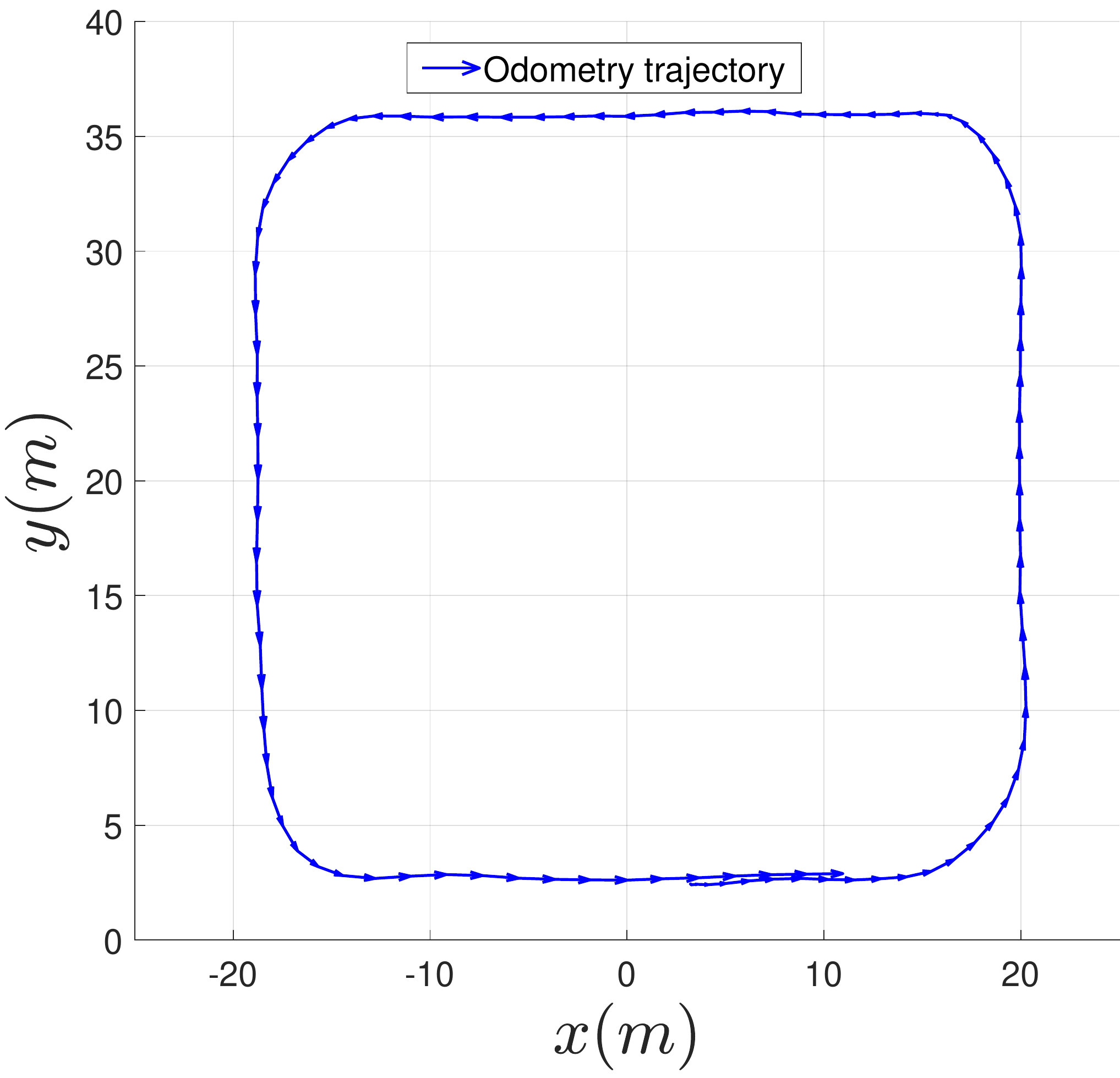}
\caption{Position data for perimeter monitoring task}
\label{fig:PM}
\end{figure}

 \begin{figure*}[ht]
	\begin{subfigure}[t]{0.5\textwidth}
		\centering
		\includegraphics[width=8cm,height=6cm]{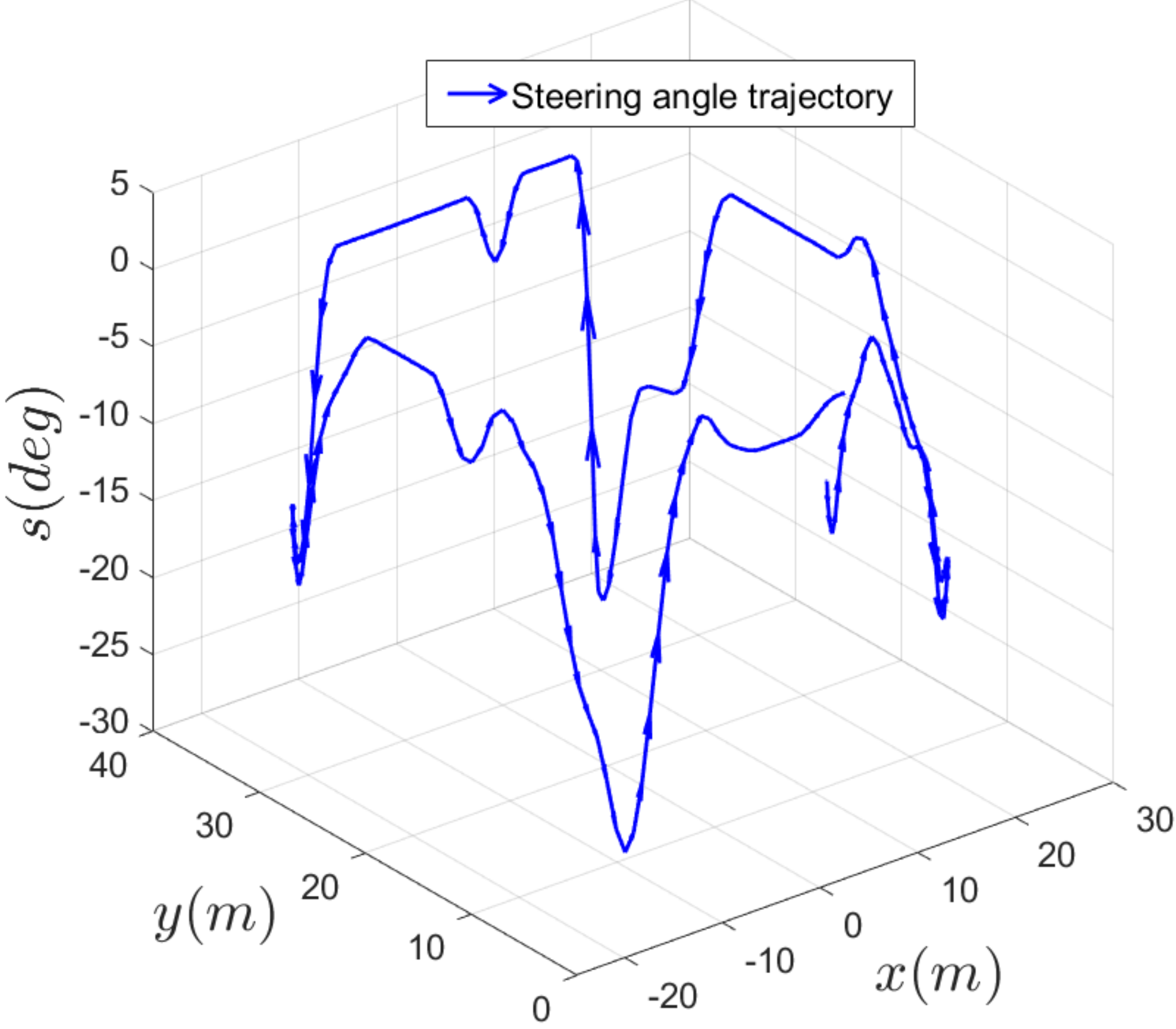}
		\caption{Steering angle w.r.t position}
		\label{fig:steering}
	\end{subfigure}%
	\begin{subfigure}[t]{0.5\textwidth}
		\centering
		\includegraphics[width=9cm,height=6cm]{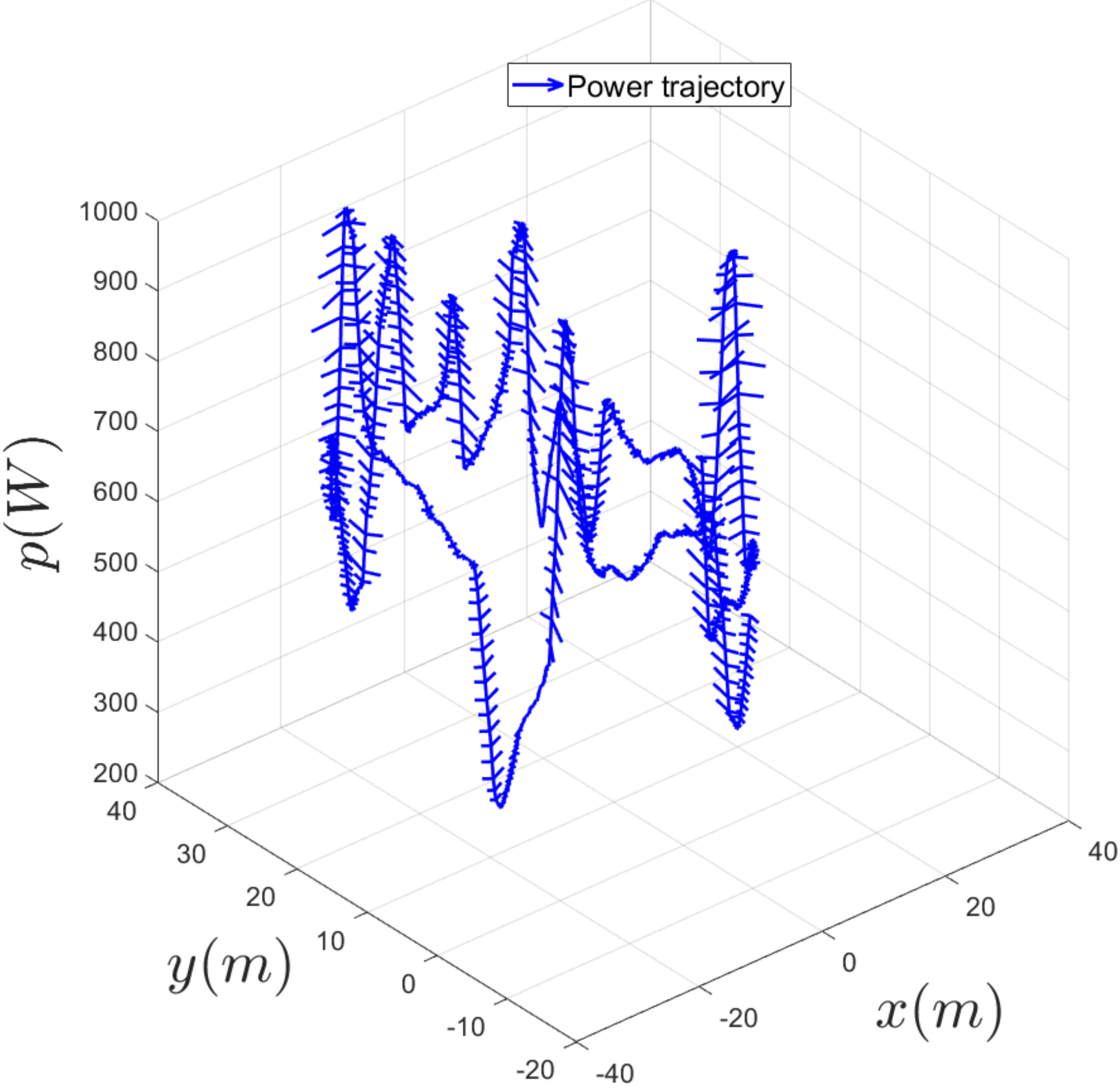}
		\caption{Power w.r.t position}
		\label{fig:power}
	\end{subfigure}
	\caption{Control variables plotted w.r.t position}
	\label{fig:control}
\end{figure*}

Fig. \ref{fig:PM} plots the one lap (from the four laps) odometry ($x$ and $y$ positions) data for the perimeter monitoring task. Fig.\ref{fig:control} shows the steering angle w.r.t the vehicle's position (Fig. \ref{fig:steering}) and the rotor power w.r.t the vehicle's position (Fig. \ref{fig:power}). For simplicity, in Fig \ref{fig:control}, plotted only the data from one lap (i.e., about 800 data points).

To test the reliability and quantify the expected delay of the data exchange between the two vehicles, we have used the ONE simulator \cite{keranen2009one}. It is a network simulator designed for testing communications among moving objects. Moreover, we have analyzed how the packet loss and delay affect the proposed learned DBN models for abnormality detection.

Considering the current state-of-the-art, the IEEE 802.11p protocol is one of the most feasible and widely considered in the inter-vehicles communication scenario, especially in autonomous vehicle networks \cite{vegni2015survey}.

A new interface has been created in the ONE simulator in order to be able to model the inter-vehicle channel as a Rician channel and to set different values for its parameters, including transmitted power, central frequency, and Rician K factor.
 
We assumed that the data to be communicated between the ego-things are: XY position, steering angle ($s$), and power ($p$) of the rotor of the iCab vehicles together with a time stamp. In this way, we assume that the amount of data to be sent is 4 Bytes for the position + 2 Bytes for the steering angle + 2 Bytes for the rotor power + 4 Bytes for the time stamp. The total data payload size is 12 Bytes. Assuming UDP, IP, and IEEE 802.11p as transport, network, and data link layer protocols, respectively, the overall size of each data packet is 12 + 8 + 20 + 28 + 6 = 74 Bytes.

\section{Results} \label{sec:results}

Results of offline learning of DBN models are here not described  in all their steps but just providing  some general overview. Then, the application of models learned in the online test phase is described in more detail to highlight their application to the agents of the ego-thing network.
\subsection{Offline training phase}

In order to collect training and test data sets, the vehicles performed platooning operation four times one after another. The size of each data sequence is about 3200 (i.e. 800 samples per each).
The null force filter \cite{campo2017static} with a single switching variable corresponds to a unique dynamic model that assumes no state change produces error sequence when the data sequence violates from this rule. The DBN model in the discrete level (i.e. word level) has been learned separately for each ego-thing while they were doing the same co-operative task. The initial filter \cite{campo2017static} has been applied to all the agents in the network.  The error sequence produced by this initial filter has been clustered to define state-dependent linear dynamic models characterizing the state as varying according to average derivatives and their covariance. 

The total number of clusters (nodes) obtained by clustering the states and errors (obtained from the initial filter) were $35$ for state space and also $35$ clusters for state derivatives with corresponding dynamic models. The GNG reached convergence in this number. Each node cluster corresponds to a letter with respect to respective vocabularies, and the word list has been composed by the different possible combinations of letters from state and state derivative vocabularies (i.e. switching variables and related dynamic models). Then a unique label is given to each letter combinations, and at last, $442$ such unique combinations of letters (i.e. words) were kept. A transition matrix at the discrete level was then estimated for each agent whose size was $442X442$. This information constitutes the DBN model of the MJPF filter to be applied to each agent.

\subsection{Online test phase}

\begin{figure*}[ht]
\centering
 	\includegraphics[width=17cm,height=3.5cm]{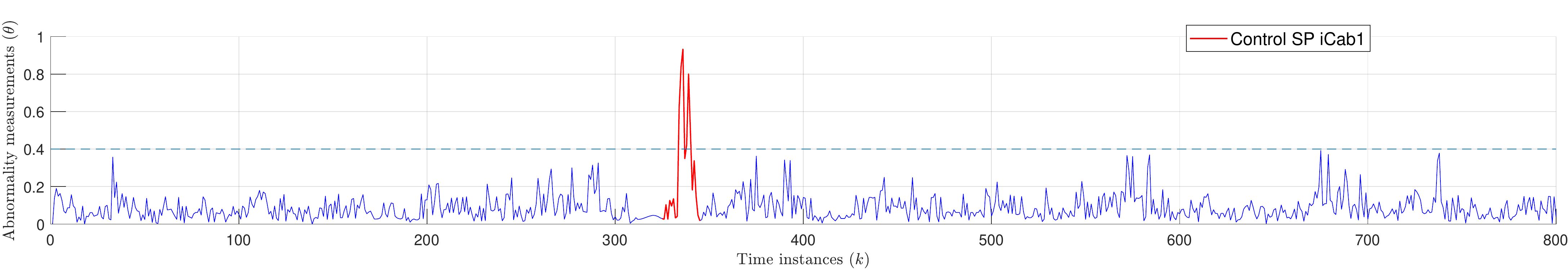}
\caption{Abnormality measurements plot for iCab 1}
\label{fig:abnicab1}
\end{figure*}

\begin{figure*}[ht]
\centering
 	\includegraphics[width=17cm,height=3.5cm]{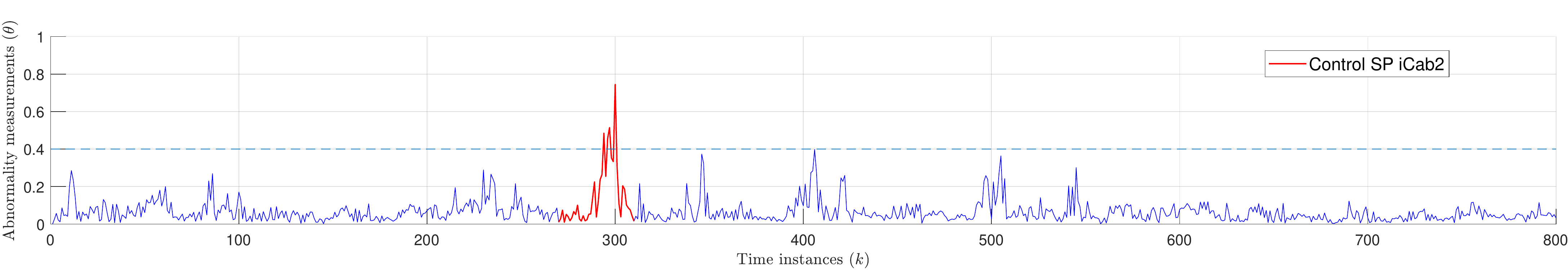}
\caption{Abnormality measurements plot for iCab 2}
\label{fig:abnicab2}
\end{figure*}

In the online test phase, the data set of scenario II, i.e. emergency stop (refer Sec. \ref{1}), has been employed to check the prediction capability of switching DBN models learned in the training phase and to detect the presence of abnormal situations in the environment. The model was able to detect the abnormality situation due to the emergency brake obtaining high values of the Hellinger distance metric, as shown by the red part in Fig. \ref{fig:abnicab1} (iCab1 - leader) and \ref{fig:abnicab2} (iCab2 - follower).

As can be seen in both figures, there is a significant rise in the Hellinger distance abnormality measures during the abnormality intervals. However, the abnormality peak of the follower vehicle is not as high as the leader's one. The main reason is that after the emergency stop of the leader, the follower gradually decreased its speed rather than doing an emergency brake. We set the abnormality threshold to 0.4 (indicated by the blue dotted line in Fig. \ref{fig:abnicab1} and \ref{fig:abnicab2}) considering the average Hellinger distance value of 0.2 when vehicles operate in normal conditions.

As described before, both vehicles have their own DBN model as well as the model for other vehicles. We have shown here the performance of the DBN model for the leader vehicle in the leader itself and the follower by plotting ROC curves and comparing AUC and ACC parameters. The model for the leader in the follower vehicle can predict the future states of the leader vehicle and detect if any abnormal situations present in the environment around the leader vehicle. Once the follower vehicle detects the abnormal situation of the leader, it should adapt its own behaviour by changing its future action by appropriate decisions. It should be pointed out that in this work the focus is on abnormality detection as a basic step of collective awareness, while the impact on such additional information on decision making and online learning of new actions is beyond the scope of this paper. The important factor that needs to be considered here is the effect of the communication channel over the transmitted data between the vehicles. Such transmission loss causes the performance degradation of the DBN model and consequently the abnormality detection capability as well.

The DBN model for the leader vehicle inside the follower vehicle estimates the abnormality situation of the leader after receiving real-time observed data (i.e. steering angle and power) from the leader over the wireless channel. Due to the impact of the communication channel over the transmitted data, the DBN model performance decreases, and we have investigated how it affects the capability of detecting abnormal situations. The performance measure we used in this work is ROC curve parameters such as AUC and ACC. The main factors that affect the transmission loss are the data rate of different modulation schemes, the distance between the vehicles, and the Rician $K$-factor.

The IEEE 802.11p standard operates at 5.9 GHz central frequency, offers 10 MHz bandwidth, and allows sending data with different modulations and data rates range from 3 to 27 Mbps \cite{ieee802.11p}. We have fixed a maximum communication range to 100 m, considering that the data loss is considerable and beyond a possible realistic reliability requirement if the distance is higher. We performed different tests changing the values of data rate, modulation, and $K$-factor as shown in Table \ref{802.11values}. High $K$-factor values refer to rural scenarios where the presence of obstacles, buildings, etc. has a lower impact on the achieved performance. The sensitivity column shows the minimum values of the Signal-to-Noise Ratio (SNR) at the receiver to guarantee successful data reception \cite{10.4108/eai.31-8-2017.153052}.

\begin{table}[ht]
\centering
\caption{Simulation parameters \cite{10.4108/eai.31-8-2017.153052}}
\begin{tabular}{||c c c c||} 
 \hline
  Data rate (Mbits/sec) & Modulation & Sensitivity (dBm) & K factor \\ [0.8ex] 
 \hline\hline
 3   & BPSK    &-85&   0,1.8,2.6,3 \\ 
 9&   QPSK  & -80  & 0,1.8,2.6,3 \\
 18 &16QAM & -73&  0,1.8,2.6,3\\
  27    &64QAM & -68&  0,1.8,2.6,3 \\[1ex] 
 \hline
\end{tabular}
\label{802.11values}
\end{table}

\begin{figure}[ht]
\centering
    \includegraphics[width=1\columnwidth]{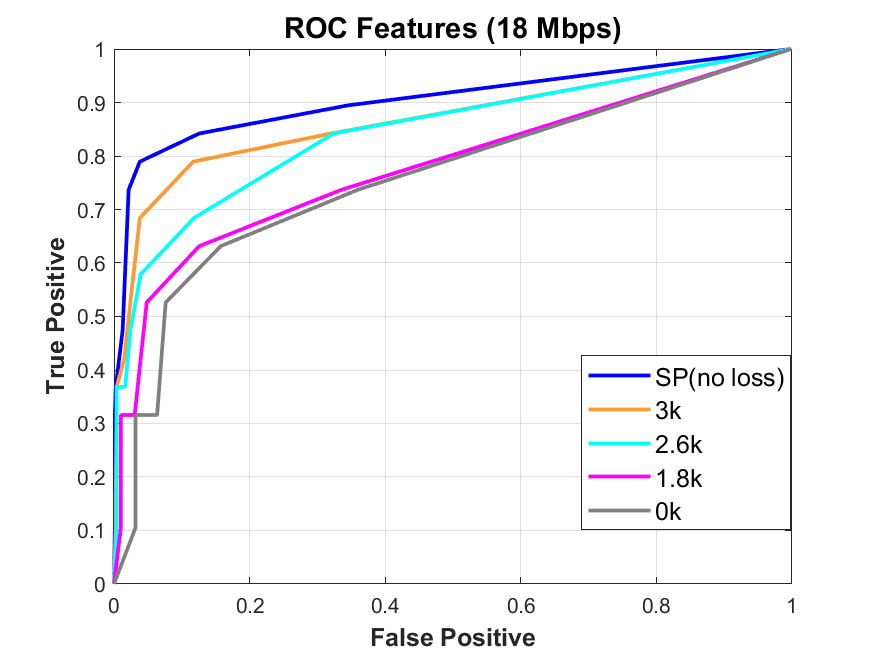}
\caption{Receiver Operating Curve (18 Mbps)}
\label{fig:ROC1}
\end{figure}

\begin{figure}[ht]
\centering
 	\includegraphics[width=1\columnwidth]{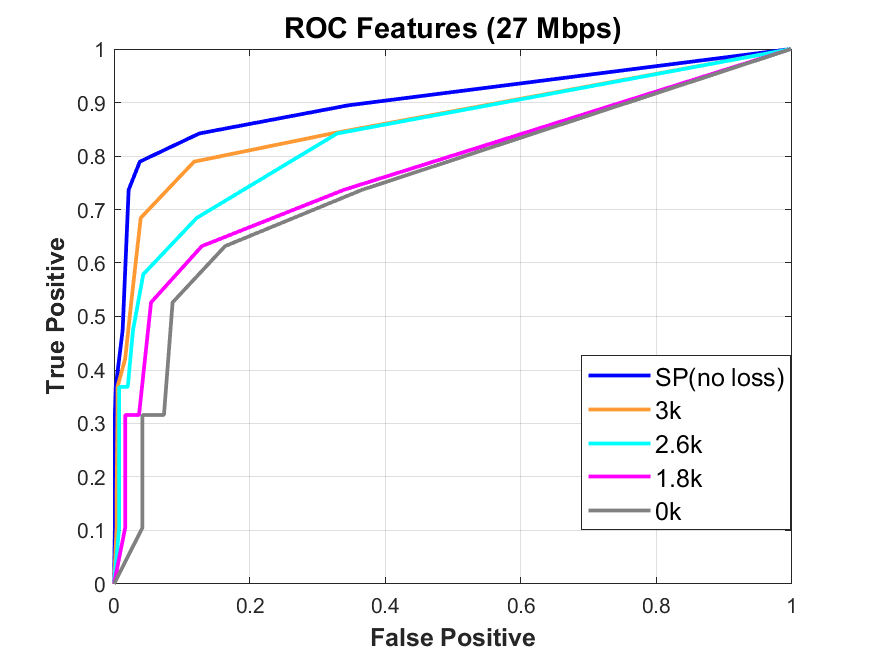}
\caption{Receiver Operating Curve (27 Mbps)}
\label{fig:ROC2}
\end{figure}

The DBN model performance in terms of ROC curve has been plotted for the leader vehicle for data rates 18 Mbps in Fig. \ref{fig:ROC1} and 27 Mbps in Fig. \ref{fig:ROC2}, respectively. These figures show the reliability of the communications in different scenarios, from completely rural ($K = 3$) to urban ($K = 0$). The blue curve refers to the case without transmission among ego-things, i.e. the ideal case of complete knowledge, and has been inserted as a comparison.

When the data rate is 18Mbps, our learned DBN model performance is not degrading much as compared to the no transmission loss case, the performance is in the acceptable range and the model well predicts the abnormal situations. The performance degradation (in terms of AUC) and the accuracy in prediction (in terms of ACC) for 18 Mbps and 27 Mbps data rates are summarized in Tables \ref{table:roc18} and \ref{table:roc27}, respectively. When the environment changes from rural to suburban to urban, the performance of the model again degrades. Finally, when in case of no line of sight component (LOS) ($K = 3$) the value of AUC and ACC in the ROC curve further reduced.

 \begin{table}[ht]
\centering
\caption{ROC features: Data rate = 18Mbps}
\begin{tabular}{||c c c ||} 
 \hline
  K-factor & Area under the curve(AUC) & Accuracy(ACC) \\ [0.8ex] 
 \hline\hline
 No loss  &\textbf{ 0.9039} &\textbf{0.9826} \\ 
 3 & 0.86665   &   0.9814  \\
 2.6 & 0.8444 & 0.9814\\
 1.8 & 0.7788 & 0.9764 \\
  0 & 0.7059 & 0.9764 \\[1ex] 

 \hline
\end{tabular}
\label{table:roc18}
\end{table}

  \begin{table}[ht]
\centering
\caption{ROC features: Data rate = 27Mbps}
\begin{tabular}{||c c c ||} 
 \hline
  K-factor & Area under the curve(AUC) & Accuracy(ACC) \\ [0.8ex] 
 \hline\hline
 No loss  &\textbf{0.9039} & \textbf{0.9826} \\ 
 3 & 0.8653   &   0.9801  \\
 2.6 & 0.8409 & 0.9777\\
  1.8 & 0.7743 & 0.9764 \\
  0 & 0.6994 & 0.9764 \\[1ex] 
 \hline
\end{tabular}
\label{table:roc27}
\end{table}

Moreover, the distance between the vehicles plays a role in packet losses. To analyze the relationship between distance, delay, and data packet loss, we focused on Scenario I but changing the velocity trace of the follower, in order to let the distance among them change during the simulation.
\begin{figure*}[ht]
\centering
\hspace*{-1.6cm} 
 	\includegraphics[width=21cm,height=3.5cm]{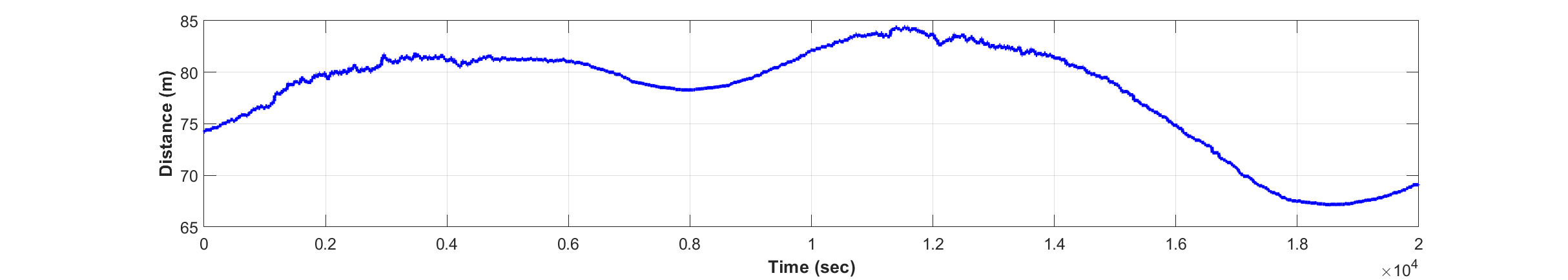}
\caption{Distance vs time}
\label{fig:distime}
\end{figure*}

\begin{figure*}[ht]
\centering
\hspace*{-1.6cm} 
 	\includegraphics[width=21cm,height=3.5cm]{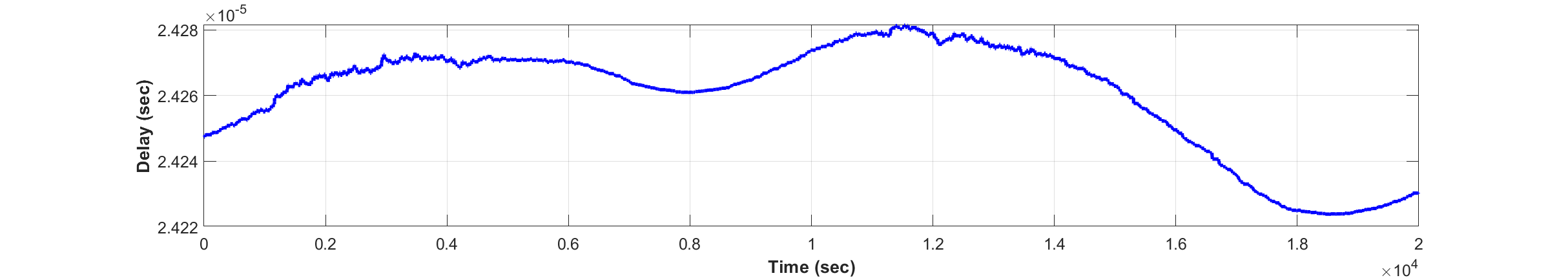}
\caption{Delay vs time}
\label{fig:delaytime}
\end{figure*}

\begin{figure*}[ht]
\centering
\hspace*{-1.5cm}
 	\includegraphics[width=21cm,height=4.5cm]{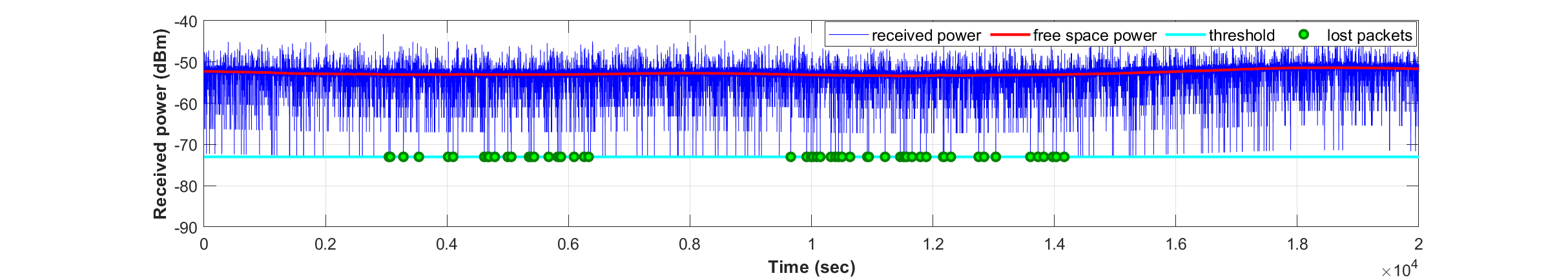}
\caption{Received power and packet losses}
\label{fig:lostpkt}
\end{figure*}

Fig. \ref{fig:distime} and \ref{fig:delaytime} show how the distance between the two vehicles and the communication delay between them change over time, respectively, while the received SNR (blue plot) and the data packet losses (green dots) are shown in Fig. \ref{fig:lostpkt}. It is evident from the figures that when the distance increases the packet loss also increases. Fig. \ref{fig:lostpkt} also shows the power in free space (red line) and sensitivity threshold (light blue line) of the power corresponds to the data rates we considered and finally the lost packets as it did not satisfy the threshold limit of the minimum received power. The overall amount of packets lost is shown in Table \ref{table:lossprob}.

\begin{table}[ht]
\centering
\caption{Packet loss ratio for different K values and different data rates}
\begin{tabular}{||c c c ||} 
 \hline
  K-factor & Packet loss ratio (18 Mbps) & Packet loss ratio (27 Mbps) \\ [0.8ex] 
 \hline\hline
 3 & 0.0025   &   0.0037  \\
 2.6 & 0.0037 & 0.0074\\
 1.8 & 0.0099 & 0.0160 \\
 0 & 0.0310 & 0.0410 \\[1ex] 
 \hline
\end{tabular}
\label{table:lossprob}
\end{table}

Considering the shown results, different considerations can be pointed out about which data ego-things should exchange to increase the reliability of the described system. For example, if the current distance between vehicles allows obtaining a packet loss ratio below a certain threshold, the vehicles can decide to communicate the ground truth observations to the other vehicles in the network to better detect if there are any abnormalities in the environment. Otherwise, if the vehicle approaches the border of the transmission range or the distance between them exceed a certain threshold, it would be more appropriate to communicate only the abnormality measurements as soon as it detected rather than communicating all the ground truth observations. The transmission loss is directly proportional to the distance, such that if we transmit more data the loss also increases. To reduce the impact of false alarm or missed detection in sensing the abnormal situation, in such situations (higher distances), 
communicate only abnormality measurements could be more appropriate to give an indication to other ego-things in the network. Although in small distances it is recommended to exchange the observed data itself to detect abnormalities with an acceptable delay.

\section{Conclusion and future work} \label{sec:conclusions}

This paper proposed a method to develop a collective awareness (CA) model and to recognize abnormal situations in smart object networks. Each entity learns a collective awareness model (i.e., a set of Dynamic Bayesian Network (DBN) models) describing the normal behaviour of itself and all the other entities in the network. The considered abnormality metric is based on the Hellinger distance between the predicted states by the learned DBN models and the real-time ground truth observations. A Markov Jump Particle Filter (MJPF) is employed to infer the future states of the entities. The abnormality metric values calculated in each of the DBN models suggest that our method provides good performance in detecting the environmental abnormalities. Moreover, information exchange among entities has been considered in order to enhance the proposed strategy.

The considered test scenario is composed of two smart vehicles, one (the follower) following the other (the leader), which move along a predefined track. Communication performance has been collected in order to verify the reliability of the data exchange, quantify the expected performance in terms of delay and loss and consider how these performances could affect the abnormality detection process. We investigated the DBN model performance in the case where each object communicates the ground truth observations to the other entities in the network. To compare the performance with different parameters of the considered channel model (Rician model), such as $K$-factor, distance, and data rates, we have plotted ROC curves and calculated reliability (Area Under the Curve - AUC) and accuracy (ACC) metrics.

In the future, this work can be extended by learning new DBN models whenever the entities pass through the new experiences. Another direction could be optimizing the model in a way that the same model could be used for all the objects in the network only by changing specific parameters. Moreover, the design of a  decision making module capable to use  abnormality situations with respect to available DBN models in order to adapt its own  actions in unknown scenarios  (by considering different test scenarios) is  a topic of  future directions of the research.

\section*{Acknowledgment}

Supported by SEGVAUTO 4.0 P2018/EMT-4362) and CICYT projects (TRA2015-63708-R and TRA2016-78886-C3-1-R).

\bibliographystyle{IEEEtran}
\bibliography{References}

\begin{thebibliography}{10}
\providecommand{\url}[1]{#1}
\csname url@samestyle\endcsname
\providecommand{\newblock}{\relax}
\providecommand{\bibinfo}[2]{#2}
\providecommand{\BIBentrySTDinterwordspacing}{\spaceskip=0pt\relax}
\providecommand{\BIBentryALTinterwordstretchfactor}{4}
\providecommand{\BIBentryALTinterwordspacing}{\spaceskip=\fontdimen2\font plus
\BIBentryALTinterwordstretchfactor\fontdimen3\font minus
  \fontdimen4\font\relax}
\providecommand{\BIBforeignlanguage}[2]{{%
\expandafter\ifx\csname l@#1\endcsname\relax
\typeout{** WARNING: IEEEtran.bst: No hyphenation pattern has been}%
\typeout{** loaded for the language `#1'. Using the pattern for}%
\typeout{** the default language instead.}%
\else
\language=\csname l@#1\endcsname
\fi
#2}}
\providecommand{\BIBdecl}{\relax}
\BIBdecl

\bibitem{ding2018amateur}
G.~Ding, Q.~Wu, L.~Zhang, Y.~Lin, T.~A. Tsiftsis, and Y.-D. Yao, ``An amateur
  drone surveillance system based on the cognitive internet of things,''
  \emph{IEEE Communications Magazine}, vol.~56, no.~1, pp. 29--35, 2018.

\bibitem{guerrero2015integration}
J.~A. Guerrero-Ibanez, S.~Zeadally, and J.~Contreras-Castillo, ``Integration
  challenges of intelligent transportation systems with connected vehicle,
  cloud computing, and internet of things technologies,'' \emph{IEEE Wireless
  Communications}, vol.~22, no.~6, pp. 122--128, 2015.

\bibitem{gubbi2013internet}
J.~Gubbi, R.~Buyya, S.~Marusic, and M.~Palaniswami, ``Internet of things (iot):
  A vision, architectural elements, and future directions,'' \emph{Future
  generation computer systems}, vol.~29, no.~7, pp. 1645--1660, 2013.

\bibitem{gerla2014internet}
M.~Gerla, E.-K. Lee, G.~Pau, and U.~Lee, ``Internet of vehicles: From
  intelligent grid to autonomous cars and vehicular clouds,'' in \emph{2014
  IEEE world forum on internet of things (WF-IoT)}.\hskip 1em plus 0.5em minus
  0.4em\relax IEEE, 2014, pp. 241--246.

\bibitem{murphy2002dynamic}
K.~P. Murphy and S.~Russell, ``Dynamic bayesian networks: representation,
  inference and learning,'' 2002.

\bibitem{fox2009nonparametric}
E.~Fox, E.~B. Sudderth, M.~I. Jordan, and A.~S. Willsky, ``Nonparametric
  bayesian learning of switching linear dynamical systems,'' in \emph{Advances
  in Neural Information Processing Systems}, 2009, pp. 457--464.

\bibitem{costa2006discrete}
O.~L.~V. Costa, M.~D. Fragoso, and R.~P. Marques, \emph{Discrete-time Markov
  jump linear systems}.\hskip 1em plus 0.5em minus 0.4em\relax Springer Science
  \& Business Media, 2006.

\bibitem{sarkka2007rao}
S.~S{\"a}rkk{\"a}, A.~Vehtari, and J.~Lampinen, ``Rao-blackwellized particle
  filter for multiple target tracking,'' \emph{Information Fusion}, vol.~8,
  no.~1, pp. 2--15, 2007.

\bibitem{statista}
\emph{Internet of Things (IoT) connected devices installed base worldwide from
  2015 to 2025 (in billions)},
  \url{https://www.statista.com/statistics/471264/iot-number-of-connected-devices-worldwide/}.

\bibitem{dua2014systematic}
A.~Dua, N.~Kumar, and S.~Bawa, ``A systematic review on routing protocols for
  vehicular ad hoc networks,'' \emph{Vehicular Communications}, vol.~1, no.~1,
  pp. 33--52, 2014.

\bibitem{baydoun2018multi}
M.~Baydoun, M.~Ravanbakhsh, D.~Campo, P.~Marin, D.~Martin, L.~Marcenaro,
  A.~Cavallaro, and C.~S. Regazzoni, ``A multi-perspective approach to anomaly
  detection for self-aware embodied agents,'' \emph{arXiv preprint
  arXiv:1803.06579}, 2018.

\bibitem{ravanbakhsh2018learning}
M.~Ravanbakhsh, M.~Baydoun, D.~Campo, P.~Marin, D.~Martin, L.~Marcenaro, and
  C.~S. Regazzoni, ``Learning multi-modal self-awareness models for autonomous
  vehicles from human driving,'' \emph{arXiv preprint arXiv:1806.02609}, 2018.

\bibitem{6823640}
N.~{Lu}, N.~{Cheng}, N.~{Zhang}, X.~{Shen}, and J.~W. {Mark}, ``Connected
  vehicles: Solutions and challenges,'' \emph{IEEE Internet of Things Journal},
  vol.~1, no.~4, pp. 289--299, Aug 2014.

\bibitem{parno2005challenges}
B.~Parno and A.~Perrig, ``Challenges in securing vehicular networks,'' in
  \emph{Workshop on hot topics in networks (HotNets-IV)}.\hskip 1em plus 0.5em
  minus 0.4em\relax Maryland, USA, 2005, pp. 1--6.

\bibitem{morin2006levels}
A.~Morin, ``Levels of consciousness and self-awareness: A comparison and
  integration of various neurocognitive views,'' \emph{Consciousness and
  cognition}, vol.~15, no.~2, pp. 358--371, 2006.

\bibitem{hood2012self}
B.~Hood, \emph{The self illusion: How the social brain creates identity}.\hskip
  1em plus 0.5em minus 0.4em\relax Oxford University Press, 2012.

\bibitem{koch2016neural}
C.~Koch, M.~Massimini, M.~Boly, and G.~Tononi, ``Neural correlates of
  consciousness: progress and problems,'' \emph{Nature Reviews Neuroscience},
  vol.~17, no.~5, p. 307, 2016.

\bibitem{chella2009machine}
A.~Chella and R.~Manzotti, ``Machine consciousness: a manifesto for robotics,''
  \emph{International Journal of Machine Consciousness}, vol.~1, no.~01, pp.
  33--51, 2009.

\bibitem{lewis2011survey}
P.~R. Lewis, A.~Chandra, S.~Parsons, E.~Robinson, K.~Glette, R.~Bahsoon,
  J.~Torresen, and X.~Yao, ``A survey of self-awareness and its application in
  computing systems,'' in \emph{Self-Adaptive and Self-Organizing Systems
  Workshops (SASOW), 2011 Fifth IEEE Conference on}.\hskip 1em plus 0.5em minus
  0.4em\relax IEEE, 2011, pp. 102--107.

\bibitem{haggard2009experience}
P.~Haggard and M.~Tsakiris, ``The experience of agency: Feelings, judgments,
  and responsibility,'' \emph{Current Directions in Psychological Science},
  vol.~18, no.~4, pp. 242--246, 2009.

\bibitem{dehaene2001towards}
S.~Dehaene and L.~Naccache, ``Towards a cognitive neuroscience of
  consciousness: basic evidence and a workspace framework,'' \emph{Cognition},
  vol.~79, no. 1-2, pp. 1--37, 2001.

\bibitem{shadlen2011consciousness}
M.~N. Shadlen and R.~Kiani, ``Consciousness as a decision to engage,'' in
  \emph{Characterizing consciousness: from cognition to the clinic?}\hskip 1em
  plus 0.5em minus 0.4em\relax Springer, 2011, pp. 27--46.

\bibitem{seth2012interoceptive}
A.~K. Seth, K.~Suzuki, and H.~D. Critchley, ``An interoceptive predictive
  coding model of conscious presence,'' \emph{Frontiers in psychology}, vol.~2,
  p. 395, 2012.

\bibitem{han2017tdoa}
S.-K. Han, W.-S. Ra, and J.~B. Park, ``Tdoa/fdoa based target tracking with
  imperfect position and velocity data of distributed moving sensors,''
  \emph{International Journal of Control, Automation and Systems}, vol.~15,
  no.~3, pp. 1155--1166, 2017.

\bibitem{campo2017static}
D.~Campo, A.~Betancourt, L.~Marcenaro, and C.~Regazzoni, ``Static force field
  representation of environments based on agents’ nonlinear motions,''
  \emph{EURASIP Journal on Advances in Signal Processing}, vol. 2017, no.~1,
  p.~13, 2017.

\bibitem{Friston1211}
K.~Friston and S.~Kiebel, ``Predictive coding under the free-energy
  principle,'' \emph{Philosophical Transactions of the Royal Society of London
  B: Biological Sciences}, vol. 364, no. 1521, pp. 1211--1221, 2009.

\bibitem{balaji2011bayesian}
B.~Balaji and K.~Friston, ``Bayesian state estimation using generalized
  coordinates,'' in \emph{Signal Processing, Sensor Fusion, and Target
  Recognition XX}, vol. 8050.\hskip 1em plus 0.5em minus 0.4em\relax
  International Society for Optics and Photonics, 2011, p. 80501Y.

\bibitem{Koller2001SamplingIF}
D.~Koller and U.~Lerner, ``Sampling in factored dynamic systems,'' in
  \emph{Sequential Monte Carlo Methods in Practice}, 2001.

\bibitem{Fritzke:1994:GNG:2998687.2998765}
\BIBentryALTinterwordspacing
B.~Fritzke, ``A growing neural gas network learns topologies,'' in
  \emph{Proceedings of the 7th International Conference on Neural Information
  Processing Systems}, ser. NIPS'94.\hskip 1em plus 0.5em minus 0.4em\relax
  Cambridge, MA, USA: MIT Press, 1994, pp. 625--632. [Online]. Available:
  \url{http://dl.acm.org/citation.cfm?id=2998687.2998765}
\BIBentrySTDinterwordspacing

\bibitem{macqueen1967some}
J.~MacQueen \emph{et~al.}, ``Some methods for classification and analysis of
  multivariate observations,'' in \emph{Proceedings of the fifth Berkeley
  symposium on mathematical statistics and probability}, vol.~1, no.~14.\hskip
  1em plus 0.5em minus 0.4em\relax Oakland, CA, USA, 1967, pp. 281--297.

\bibitem{kohonen1982self}
T.~Kohonen, ``Self-organized formation of topologically correct feature maps,''
  \emph{Biological cybernetics}, vol.~43, no.~1, pp. 59--69, 1982.

\bibitem{fritzke1995growing}
B.~Fritzke, ``A growing neural gas network learns topologies,'' in
  \emph{Advances in neural information processing systems}, 1995, pp. 625--632.

\bibitem{fritzke1997self}
------, ``A self-organizing network that can follow non-stationary
  distributions,'' in \emph{International conference on artificial neural
  networks}.\hskip 1em plus 0.5em minus 0.4em\relax Springer, 1997, pp.
  613--618.

\bibitem{baydoun2018learning}
M.~Baydoun, D.~Campo, V.~Sanguineti, L.~Marcenaro, A.~Cavallaro, and
  C.~Regazzoni, ``Learning switching models for abnormality detection for
  autonomous driving,'' in \emph{2018 21st International Conference on
  Information Fusion (FUSION)}.\hskip 1em plus 0.5em minus 0.4em\relax IEEE,
  2018, pp. 2606--2613.

\bibitem{8767204}
D.~{Kanapram}, D.~{Campo}, M.~{Baydoun}, L.~{Marcenaro}, E.~L. {Bodanese},
  C.~{Regazzoni}, and M.~{Marchese}, ``Dynamic bayesian approach for
  decision-making in ego-things,'' in \emph{2019 IEEE 5th World Forum on
  Internet of Things (WF-IoT)}, April 2019, pp. 909--914.

\bibitem{welch1995introduction}
G.~Welch, G.~Bishop \emph{et~al.}, ``An introduction to the kalman filter,''
  1995.

\bibitem{210672}
N.~J. {Gordon}, D.~J. {Salmond}, and A.~F.~M. {Smith}, ``Novel approach to
  nonlinear/non-gaussian bayesian state estimation,'' \emph{IEE Proceedings F -
  Radar and Signal Processing}, vol. 140, no.~2, pp. 107--113, April 1993.

\bibitem{yang2005particle}
N.~Yang, W.~F. Tian, Z.~H. Jin, and C.~B. Zhang, ``Particle filter for sensor
  fusion in a land vehicle navigation system,'' \emph{Measurement science and
  technology}, vol.~16, no.~3, p. 677, 2005.

\bibitem{beran1977minimum}
R.~Beran \emph{et~al.}, ``Minimum hellinger distance estimates for parametric
  models,'' \emph{The annals of Statistics}, vol.~5, no.~3, pp. 445--463, 1977.

\bibitem{bhattacharyya1943measure}
A.~Bhattacharyya, ``On a measure of divergence between two statistical
  populations defined by their probability distributions,'' \emph{Bull.
  Calcutta Math. Soc.}, vol.~35, pp. 99--109, 1943.

\bibitem{verdu2014total}
S.~Verd{\'u}, ``Total variation distance and the distribution of relative
  information,'' in \emph{2014 Information Theory and Applications Workshop
  (ITA)}.\hskip 1em plus 0.5em minus 0.4em\relax IEEE, 2014, pp. 1--3.

\bibitem{abdi2007encyclopedia}
H.~Abdi and N.~J. Salkind, ``Encyclopedia of measurement and statistics,''
  \emph{Thousand Oaks, CA: Sage. Agresti, A.(1990) Categorical data analysis.,
  New York: Wiley. Agresti, A.(1992) A survey of exact inference for
  contingency tables. Statist Sci}, vol.~7, pp. 131--153, 2007.

\bibitem{lourenzutti2014hellinger}
R.~Lourenzutti and R.~A. Krohling, ``The hellinger distance in multicriteria
  decision making: An illustration to the topsis and todim methods,''
  \emph{Expert Systems with Applications}, vol.~41, no.~9, pp. 4414--4421,
  2014.

\bibitem{Bhattacharyya1943}
A.~Bhattacharyya, ``{On a measure of divergence between two statistical
  populations defined by their probability distributions},'' \emph{Bulletin of
  the Calcutta Mathematical Society}, vol.~35, pp. 99--109, 1943.

\bibitem{hanley1982meaning}
J.~A. Hanley and B.~J. McNeil, ``The meaning and use of the area under a
  receiver operating characteristic (roc) curve.'' \emph{Radiology}, vol. 143,
  no.~1, pp. 29--36, 1982.

\bibitem{6802414}
S.~{Zhu}, T.~S. {Ghazaany}, S.~M.~R. {Jones}, R.~A. {Abd-Alhameed}, J.~M.
  {Noras}, T.~{Van Buren}, J.~{Wilson}, T.~{Suggett}, and S.~{Marker},
  ``Probability distribution of rician $k$-factor in urban, suburban and rural
  areas using real-world captured data,'' \emph{IEEE Transactions on Antennas
  and Propagation}, vol.~62, no.~7, pp. 3835--3839, July 2014.

\bibitem{goldsmith2005wireless}
A.~Goldsmith, \emph{Wireless communications}.\hskip 1em plus 0.5em minus
  0.4em\relax Cambridge university press, 2005.

\bibitem{marin2018global}
P.~Marin-Plaza, A.~Hussein, D.~Martin, and A.~d.~l. Escalera, ``Global and
  local path planning study in a ros-based research platform for autonomous
  vehicles,'' \emph{Journal of Advanced Transportation}, vol. 2018, 2018.

\bibitem{keranen2009one}
A.~Ker{\"a}nen, J.~Ott, and T.~K{\"a}rkk{\"a}inen, ``The one simulator for dtn
  protocol evaluation,'' in \emph{Proceedings of the 2nd international
  conference on simulation tools and techniques}.\hskip 1em plus 0.5em minus
  0.4em\relax ICST (Institute for Computer Sciences, Social-Informatics
  and~…, 2009, p.~55.

\bibitem{vegni2015survey}
A.~M. Vegni and V.~Loscri, ``A survey on vehicular social networks,''
  \emph{IEEE Communications Surveys \& Tutorials}, vol.~17, no.~4, pp.
  2397--2419, 2015.

\bibitem{ieee802.11p}
``Ieee standard for information technology-- local and metropolitan area
  networks-- specific requirements-- part 11: Wireless lan medium access
  control (mac) and physical layer (phy) specifications amendment 6: Wireless
  access in vehicular environments,'' \emph{IEEE Std 802.11p-2010 (Amendment to
  IEEE Std 802.11-2007 as amended by IEEE Std 802.11k-2008, IEEE Std
  802.11r-2008, IEEE Std 802.11y-2008, IEEE Std 802.11n-2009, and IEEE Std
  802.11w-2009)}, pp. 1--51, July 2010.

\bibitem{10.4108/eai.31-8-2017.153052}
A.~Bazzi, B.~M. Masini, and A.~Zanella, ``Cooperative awareness in the internet
  of vehicles for safety enhancement,'' \emph{EAI Endorsed Transactions on
  Internet of Things}, vol.~3, no.~9, 8 2017.

\end{thebibliography}

\end{document}